\documentclass[10pt,twocolumn,letterpaper]{article}

\usepackage{cvpr}
\usepackage{times}
\usepackage{graphicx}
\usepackage{amsmath,amssymb,amsthm}

\usepackage{subfig}
\usepackage{enumitem}

\newcommand{\argmax}{\operatorname*{argmax}}
\newcommand{\argmin}{\operatorname*{argmin}}
\usepackage{algorithm,algorithmic}
\newcommand{\D}{\mathcal{D}}
\renewcommand{\P}{\mathcal{P}}
\newcommand{\X}{\mathcal{X}}

\newcommand{\N}{\mathbb{N}}
\newcommand{\R}{\mathbb{R}}

\usepackage[pagebackref=true,breaklinks=true,letterpaper=true,colorlinks,bookmarks=false]{hyperref}

\cvprfinalcopy

\begin{document}

%%%%%%%%% TITLE
\title{iCaRL: Incremental Classifier and Representation Learning}

\author{Sylvestre-Alvise Rebuffi\\
University of Oxford/IST Austria\\
\and
Alexander Kolesnikov, \ Georg Sperl, \ Christoph H. Lampert\\
IST Austria
}

\maketitle

%%%%%%%%% ABSTRACT
\begin{abstract}
A major open problem on the road to artificial intelligence 
is the development of incrementally learning systems that 
learn about more and more concepts over time from a stream 
of data.
In this work, we introduce a new training strategy, iCaRL, 
that allows learning in such a class-incremental way: 
only the training data for a small number of classes has 
to be present at the same time and new classes can be 
added progressively.

iCaRL learns strong classifiers and a data representation
simultaneously. This distinguishes it from earlier works that
were fundamentally limited to fixed data representations and
therefore incompatible with deep learning architectures.
We show by experiments on CIFAR-100 and ImageNet ILSVRC~2012 data 
that iCaRL can learn many classes incrementally over
a long period of time where other strategies quickly fail. 
\end{abstract}

%%%%%%%%%%%%%%%%%%%%%%%%%%%%%%%%%%%%%%%%%%%%%%%%%%%%%%%%%%%%%%
%%%%%%%%%%%%%%%%%%%%%%%%%%%%%%%%%%%%%%%%%%%%%%%%%%%%%%%%%%%%%%
%%%%%%%%%%%\input{introduction.tex}
%%%%%%%%%%%%%%%%%%%%%%%%%%%%%%%%%%%%%%%%%%%%%%%%%%%%%%%%%%%%%%
%%%%%%%%%%%%%%%%%%%%%%%%%%%%%%%%%%%%%%%%%%%%%%%%%%%%%%%%%%%%%%

\section{Introduction}\label{sec:intro}
Natural vision systems are inherently incremental: new visual 
information is continuously incorporated while existing 
knowledge is preserved. 
For example, a child visiting the zoo will learn about many 
new animals without forgetting the pet it has at home.
In contrast, most artificial
object recognition systems can only be trained in a 
batch setting, 
where all object classes are known in advance and they 
the training data of all classes can be accessed at 
the same time and in arbitrary order. 

As the field of computer vision moves closer towards 
artificial intelligence it becomes apparent that more 
flexible strategies are required to handle the large-scale 
and dynamic properties of real-world object categorization 
situations. 
At the very least, a visual object classification system
should be able to incrementally learn about new classes, 
when training data for them becomes available. 
We call this scenario \emph{class-incremental learning}.

\begin{figure}[t]\centering
\includegraphics[width=\columnwidth]{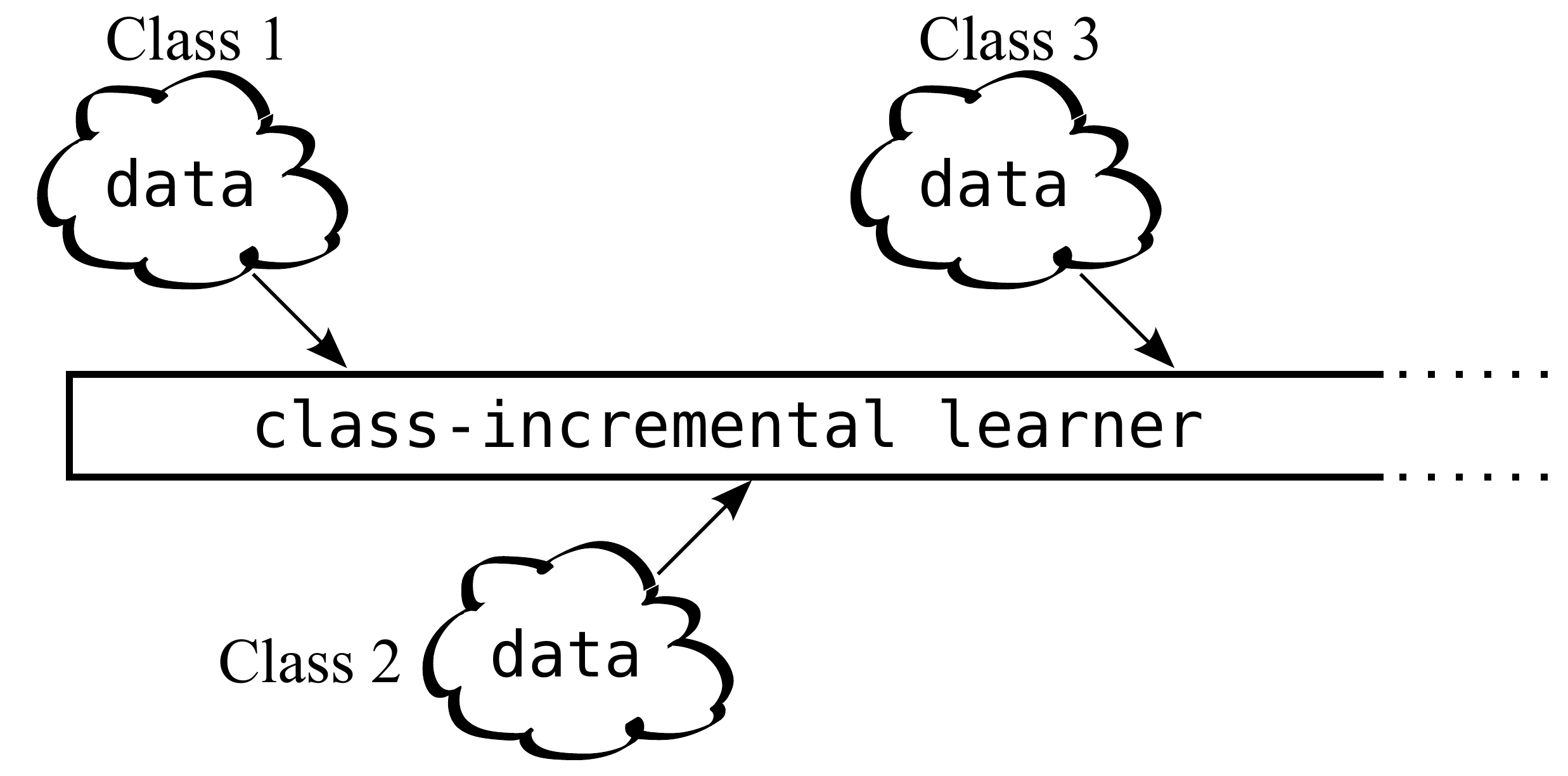}
\caption{Class-incremental learning: an algorithm learns 
continuously from a sequential data stream in which new 
classes occur. At any time, the learner is able 
to perform multi-class classification for all 
classes observed so far. 
}
\end{figure}

Formally, we demand the following three properties 
of an algorithm to qualify as class-incremental: 
\begin{itemize}[itemsep=0pt,topsep=2pt]
\item[i)] it should be trainable from a stream of data in 
which examples of different classes occur at different times,
\item[ii)] it should at any time provide a competitive 
multi-class classifier for the classes observed so far, 
\item[iii)] its computational requirements and memory 
footprint should remain bounded, or at least grow very slowly, 
with respect to the number of classes seen so far.
\end{itemize}
The first two criteria express the essence of class-incremental 
learning. The third criterion prevents trivial algorithms, 
such as storing all training examples and retraining an ordinary 
multi-class classifier whenever new data becomes available. 

Interestingly, despite the vast progress that image classification 
has made over the last decades, there is not a single satisfactory 
class-incremental learning algorithm these days. 
Most existing multi-class techniques simply violate $i)$ or $ii)$ 
as they can only handle a fixed number of classes and/or 
need all training data to be available at the same time. 
Naively, one could try to overcome this by training classifiers 
from class-incremental data streams, \eg using stochastic gradient 
descent optimization. This, however, will cause the classification 
accuracy to quickly deteriorate, an effect known in the literature 
as \emph{catastrophic forgetting} or 
\emph{catastrophic interference}~\cite{mccloskey1989catastrophic}. 
The few existing techniques that do fulfill the above properties are 
principally limited to situations with a fixed data representation. 
They cannot be extended to deep architectures that learn classifiers 
and feature representations at the same time and are therefore not 
competitive anymore in terms of classification accuracy. 
More related work is discussed in Section~\ref{sec:relatedwork}.

In this work, we introduce \emph{iCaRL (incremental classifier and 
representation learning)}, a practical strategy for simultaneously 
learning classifiers and a feature representation in the 
class-incremental setting.
Based on a careful analysis of the shortcomings of existing approaches, 
we introduce three main components that in combination allow iCaRL to 
fulfill all criteria put forth above. 
These three components are:
\begin{itemize}[itemsep=0pt,topsep=2pt]
\item classification by a \emph{nearest-mean-of-exemplars} rule,
\item \emph{prioritized exemplar selection} based on herding, 
\item representation learning using \emph{knowledge distillation}
and \emph{prototype rehearsal}.
\end{itemize}
We explain the details of these steps in Section~\ref{sec:method},
and subsequently put them into the context of previous work in 
Section~\ref{sec:relatedwork}. 
In Section~\ref{sec:experiments} we report on experiments on 
the CIFAR and ImageNet datasets that show that iCaRL is able 
to class-incrementally learn over a long periods of time, 
where other methods quickly fail.
Finally, we conclude in Section~\ref{sec:conclusion} with a discussion 
of remaining limitations and future work.

%%%%%%%%%%%%%%%%%%%%%%%%%%%%%%%%%%%%%%%%%%%%%%%%%%%%%%%%%%%%%%
%%%%%%%%%%%%%%%%%%%%%%%%%%%%%%%%%%%%%%%%%%%%%%%%%%%%%%%%%%%%%%
%%%%%%%%%%%\input{method.tex}
%%%%%%%%%%%%%%%%%%%%%%%%%%%%%%%%%%%%%%%%%%%%%%%%%%%%%%%%%%%%%%
%%%%%%%%%%%%%%%%%%%%%%%%%%%%%%%%%%%%%%%%%%%%%%%%%%%%%%%%%%%%%
\renewcommand{\phi}{\varphi}
\section{Method}\label{sec:method}
In this section we describe iCaRL's main components 
and explain how their combination allows true 
class-incremental learning. 
Section~\ref{subsec:learning} explains the underlying 
architecture and gives a high-level overview of 
the training and classification steps.
Sections~\ref{subsec:classifier} to \ref{subsec:exemplars} 
then provides the algorithmic details and explains
the design choices. 

\begin{algorithm}[t]
\caption{iCaRL \textsc{Classify}}\label{alg:icarl-classify}
    \begin{algorithmic}
        \INPUT $x$ \qquad\qquad\qquad\qquad\quad \  // image to be classified
        \REQUIRE $\P=(P_1,\dots,P_t)$ \quad\ \  // class exemplar sets 
        \REQUIRE $\phi:\X\to\R^d$ \qquad\qquad // feature map 
        \FOR{$y=1,\dots,t$}
        \STATE $\displaystyle\mu_y \leftarrow \frac{1}{|P_y|}\sum_{p\in P_y} \phi(p)$ \qquad // mean-of-exemplars %\vspace{-.2\baselineskip}
        \ENDFOR
        \STATE $\displaystyle y^\ast\leftarrow \argmin\limits_{y=1,\dots,t} \| \phi(x) - \mu_y\|$ \quad // nearest prototype
        \OUTPUT class label $y^\ast$ 
    \end{algorithmic}
\end{algorithm}

\begin{algorithm}[t]
\caption{iCaRL \textsc{IncrementalTrain}}\label{alg:icarl-train}
    \begin{algorithmic}
        \INPUT $X^s,\dots,X^t$ \quad // training examples in per-class sets
        \INPUT $K$ \qquad\qquad\quad // memory size
        \REQUIRE $\Theta$  \qquad\qquad\quad // current model parameters
        \REQUIRE $\P=(P_1,\dots,P_{s-1})$ \qquad // current exemplar sets
        \smallskip
        \STATE $\Theta \leftarrow \textsc{UpdateRepresentation}(X^s,\dots,X^t; \P,\Theta)$
        \smallskip

        \STATE $m\leftarrow K/t$ \qquad // number of exemplars per class
        \FOR{$y=1,\dots,s-1$}
        \STATE $P_y \leftarrow \textsc{ReduceExemplarSet}(P_y,m)$
        \ENDFOR
        \FOR{$y=s,\dots,t$}
        \STATE $P_y \leftarrow \textsc{ConstructExemplarSet}(X_y,m,\Theta)$
        \ENDFOR
        \smallskip
        \STATE $\P \leftarrow (P_1,\dots,P_t)$   \qquad\qquad   // new exemplar sets
    \end{algorithmic}
\end{algorithm}

\subsection{Class-Incremental Classifier Learning}\label{subsec:learning}

iCaRL learns classifiers and a feature representation 
simultaneously
from on a data stream in class-incremental form, \ie sample sets $X^1,X^2,\dots$, where all examples of 
a set $X^y=\{x^y_1,\dots,x^y_{n_y}\}$ are of class $y\in\N$. 

\paragraph{Classification.} %
For classification, iCaRL relies on sets, $P_1,\dots,P_t$, 
of \emph{exemplar images} that it selects dynamically 
out of the data stream. 
There is one such exemplar set for each observed class so far, 
and iCaRL ensures that the total number of exemplar images 
never exceeds a fixed parameter $K$. 
Algorithm~\ref{alg:icarl-classify} describes the mean-of-exemplars 
classifier that is used to classify images into the set of classes 
observed so far, see Section~\ref{subsec:classifier} for a detailed 
explanation.

\paragraph{Training.} %
For training, iCaRL processes batches of classes at a time 
using an incremental learning strategy. 
Every time data for new classes is available iCaRL calls an 
update routine (Algorithm~\ref{alg:icarl-train}, see 
Sections~\ref{subsec:representation} and \ref{subsec:exemplars}). 
The routine adjusts iCaRL's \emph{internal knowledge} (the network 
parameters and exemplars) based on the additional information 
available in the \emph{new observations} (the current training data). 
This is also how iCaRL learns about the existence 
of new classes.

\paragraph{Architecture.} Under the hood, iCaRL makes use of 
a convolutional neural network (CNN)~\cite{LeCun1998}\footnote{In principle, 
the iCaRL strategy is largely architecture agnostic and could 
be use on top of other feature or metric learning strategies. 
Here, we discuss it only in the context of CNNs 
to avoid an overly general notation.}. 
We interpret the network as a \emph{trainable feature extractor}, 
$\phi:\X\to\R^d$, followed by a single classification layer with 
as many sigmoid output nodes as classes observed so far~\cite{bengio2013representation}. 
All feature vectors are $L^2$-normalized, and the results of any 
operation on feature vectors, \eg averages, are also re-normalized,
which we do not write explicitly to avoid a cluttered notation.

We denote the parameters of the network by $\Theta$, split into  
a fixed number of parameters for the feature extraction part 
and a variable number of weight vectors. We denote the latter 
by $w_1,\dots,w_t\in\R^d$, 
where here and in the following sections we use the convention 
that $t$ denotes the number of classes that have been observed so far.
The resulting network outputs are, for any class $y\in\{1,\dots,t\}$,
\begin{align}
g_y(x) = \frac{1}{1+\exp(-a_y(x))}\quad\text{with}\ \  a_y(x)=w_y^\top\!\phi(x).
\label{eq:softmax-output}
\end{align}
Note that even though one can interpret these outputs as probabilities, 
iCaRL uses the network only for representation learning, not for the actual 
classification step.

\paragraph{Resource usage.} %
Due to its incremental nature, iCaRL does not need a priori 
information about which and how many classes will occur, and 
it can --in theory-- run for an unlimited amount of time. 
At any time during its runtime its memory requirement will 
be the size of the feature extraction parameters, the 
storage of $K$ exemplar images and as many weight vectors 
as classes that have been observed.
This knowledge allows us to assign resources depending 
on the application scenario. 
If an upper bound on the number of classes is known, one 
can simply pre-allocate space for as many weight vectors 
as required and use all remaining available memory to 
store exemplars. 
Without an upper limit, one would actually grow the number of 
weight vectors over time, and decrease the size of the exemplar 
set accordingly. 
Clearly, at least one exemplar image and weight vector is 
required for each classes to be learned, so ultimately, 
only a finite number of classes can be learned, unless 
one allows for the possibility to add more resources over 
the runtime of the algorithm. 
Note that iCaRL can handle an increase of resources on-the-fly 
without retraining: it will simply not discard any exemplars 
unless it is forced to do so by memory limitations.

\subsection{Nearest-Mean-of-Exemplars Classification}\label{subsec:classifier}
iCaRL uses a \emph{nearest-mean-of-exemplars} classification strategy. 
To predict a label, $y^\ast$, for a new image, $x$, it computes a 
prototype vector for each class observed so far, $\mu_1,\dots,\mu_t$, 
where $\mu_y=\frac{1}{|P_y|}\sum_{p\in P_y} \phi(p)$ is the average 
feature vector of all exemplars for a class $y$.
It also computes the feature vector of the image that should 
be classified and assigns the class label with most similar 
prototype: 

\begin{align}
y^\ast &= \argmin\limits_{y=1,\dots,t} \| \phi(x) - \mu_y \|.
\label{eq:classification}
\end{align}

\paragraph{Background\onedot} 
The nearest-mean-of-exemplars classification rule
overcomes two major problems of the incremental learning setting, 
as can be seen by contrasting it against other possibilities for 
multi-class classification.

The usual classification rule for a neural network would be 
$y^\ast=\argmax_{y=1,\dots,t}g_y(x)$, where $g_y(x)$ is the
network output as defined in \eqref{eq:softmax-output} or 
alternatively with a softmax output layer. 
Because $\argmax_y g_y(x)=\argmax_y w_y^\top\phi(x)$, 
the network's prediction rule is equivalent to the use of 
a linear classifier with non-linear feature map $\phi$ and 
weight vectors $w_1,\dots,w_t$. 
In the class-incremental setting, it is problematic that the weight 
vectors $w_y$ are \emph{decoupled} from the feature extraction 
routine $\phi$:
whenever $\phi$ changes, all $w_1,\dots,w_t$ must be updated as well.
Otherwise, the network outputs will change uncontrollably, which is 
observable as catastrophic forgetting.
In contrast, the nearest-mean-of-exemplars rule~\eqref{eq:classification}
does not have decoupled weight vectors. The class-prototypes automatically 
change whenever the feature representation changes, making the 
classifier robust against changes of the feature 
representation. 

The choice of the average vector as prototype is inspired by 
the \emph{nearest-class-mean} classifier~\cite{mensink2013distance} 
for incremental learning with a fixed feature representation.
In the class-incremental setting, we cannot make use of the true 
class mean, since all training data would have to be stored in 
order to recompute this quantity after a representation change. 
Instead, we use the average over a flexible number of exemplars that 
are chosen in a way to provide a good approximation to the class 
mean.

Note that, because we work with normalized feature vectors, Equation~\eqref{eq:classification} 
can be written equivalently as $y^\ast = \argmax_{y} \ \mu^{\top}_y\!\phi(x)$. 
Therefore, we can also interpret the classification step as 
classification with a weight vector, but one that is not 
decoupled from the data representation but changes consistently 
with it.

\begin{algorithm}[t]
\caption{iCaRL \textsc{UpdateRepresentation}}\label{alg:representation}
    \begin{algorithmic}
    \INPUT $X^s,\dots,X^{t}$ \quad // training images of classes $s,\dots,t$ 
    \REQUIRE $\P=(P_1,\dots,P_{s-1})$ \qquad\qquad // exemplar sets 
    \REQUIRE $\Theta$ \qquad\qquad // current model parameters 
    \STATE // form combined training set:
    %\vskip-.2\baselineskip\vspace{-.25\baselineskip}
    {\small$$\D \leftarrow \!\!\!\!\bigcup_{y=s,\dots,t}\!\!\!\!\{(x,y) : x\in X^y\} 
    \ \cup\!\!\!\!\!\!\bigcup_{y=1,\dots,s-1}\!\!\!\!\!\!\{(x,y) : x\in P^y\} \!\!\!\!\!\!\!\!\!\!\!\!
    $$}
    %\vskip-.25\baselineskip\vspace{-.25\baselineskip}
    \STATE // store network outputs with pre-update parameters:
    \FOR{$y=1,\dots,s-1$}
    \STATE $q^y_i\leftarrow g_y(x_i)$ \quad for all $(x_i,\cdot)\in\D$
    \ENDFOR
    \STATE run network training (\eg BackProp) with loss function
    {\small\begin{align*}
    \ell(\Theta) \!=\! -\!\!\!\!\!\!\!\!\sum_{(x_i,y_i)\in\D}&\!\!\Big[\sum_{y=s}^{t}\!\delta_{y=y_i}\log g_{y}(x_i) \!+ \delta_{y\neq y_i}\log(1\!-\!g_{y}(x_i))
    %\quad \text{// classification loss}
    \\ 
      +&\sum_{y=1}^{s-1}\!q^y_i\log g_y(x_i) \!+ \!(1\!-\!q^y_i)\log(1\!-\!g_y(x_i))\Big]
    \end{align*}}
    %\vskip-.5\baselineskip\vspace{-.25\baselineskip}
    \STATE that consists of \emph{classification} and \emph{distillation} terms.
    \end{algorithmic}
\end{algorithm}

\subsection{Representation Learning}\label{subsec:representation}
Whenever iCaRL obtains data, $X^s,\dots,X^t$, for new classes, 
$s,\dots,t$, it updates its feature extraction routine and the exemplar set. 
Algorithm~\ref{alg:representation} lists the steps for 
incrementally improving the feature representation. 
First, iCaRL constructs an augmented training set consisting 
of the currently available training examples together with 
the stored exemplars. 
Next, the current network is evaluated for each example and 
the resulting network outputs for all previous classes are 
stored (not for the new classes, since the network has not 
been trained for these, yet).
Finally, the network parameters are updated by minimizing 
a loss function that for each new image encourages the 
network to output the correct class indicator for new 
classes \emph{(classification loss)}, and for old classes, 
to reproduce the scores stored in the previous step 
\emph{(distillation loss)}.

\paragraph{Background\onedot} 
The representation learning step resembles ordinary 
network finetuning: starting from previously learned 
network weights it minimizes a loss function over a 
training set.
As a consequence, standard end-to-end learning methods 
can be used, such as backpropagation with mini-batches, 
but also recent improvements, such as 
\emph{dropout}~\cite{SrivastavaHKSS14}, 
\emph{adaptive stepsize selection}~\cite{KingmaB14} or
\emph{batch normalization}~\cite{IoffeS15}, 
as well as potential future 
improvements. 

There are two modifications to plain finetuning that aim at
preventing or at least mitigating catastrophic forgetting.
First, the training set is augmented. It consists not only 
of the new training examples but also of the stored exemplars. 
By this it is ensured that at least some information about 
the data distribution of all previous classes enters the 
training process. 
Note that for this step it is important that the exemplars 
are stored as images, not in a feature representation that
would become outdated over time.
Second, the loss function is augmented as well.
Besides the standard classification loss, which encourages 
improvements of the feature representation that allow classifying 
the newly observed classes well, it also contains the distillation 
loss, which ensures that the discriminative information learned 
previously is not lost during the new learning step.

\begin{algorithm}[t]
\caption{iCaRL \textsc{ConstructExemplarSet}}\label{alg:newexemplars}
    \begin{algorithmic}
        \INPUT image set $X=\{x_1,\dots,x_{n}\}$ of class $y$
        \INPUT $m$ target number of exemplars
        \REQUIRE current feature function $\phi:\X\to\R^d$
        \STATE $\mu \leftarrow \frac{1}{n}\sum_{x\in X} \phi(x)$ \ \  // current class mean
        \FOR{$k=1,\dots,m$}
        \STATE $p_k \leftarrow \argmin\limits_{x\in X}
            \Big\| \mu - \frac{1}{k}[\phi(x)+\sum_{j=1}^{k-1}\phi(p_j)] \Big\|$
        \ENDFOR
        \STATE $P\leftarrow (p_1,\dots,p_{m})$
        \OUTPUT exemplar set $P$
    \end{algorithmic}
\end{algorithm}

\begin{algorithm}[t]
\caption{iCaRL \textsc{ReduceExemplarSet}}\label{alg:removeexemplars}
    \begin{algorithmic}
        \INPUT $m$ \qquad\qquad // target number of exemplars
        \INPUT $P=(p_1,\dots,p_{|P|})$ \qquad // current exemplar set
        \STATE $P\leftarrow (p_1,\dots,p_{m})$        \qquad\qquad // \ie keep only first $m$
        \OUTPUT exemplar set $P$
    \end{algorithmic}
\end{algorithm}

\subsection{Exemplar Management}\label{subsec:exemplars}

Whenever iCaRL encounters new classes it adjusts its exemplar set. 
All classes are treated equally in this, \ie, when $t$ classes have 
been observed so far and $K$ is the total number of exemplars that
can be stored, iCaRL will use $m=K/t$ exemplars (up to rounding) 
for each class. 
By this it is ensured that the available memory budget of $K$ 
exemplars is always used to full extent, but never exceeded. 

Two routines are responsible for exemplar management: one to select 
exemplars for new classes and one to reduce the sizes of the exemplar 
sets of previous classes. 
Algorithm~\ref{alg:newexemplars} describes the exemplar 
selection step. Exemplars $p_1,\dots,p_m$ are selected 
and stored iteratively until the target number, $m$, is met. 
In each step of the iteration, one more example of the 
current training set is added to the exemplar set, namely
the one that causes the average feature vector over all 
exemplars to best approximate the average feature vector 
over all training examples.
Thus, the exemplar "set" is really a prioritized list. 
The order of its elements matters, with exemplars 
earlier in the list being more important. 
The procedure for removing exemplars is specified 
in Algorithm~\ref{alg:removeexemplars}. It is 
particularly simple: 
to reduce the number of exemplars from any $m'$ to $m$, 
one discards the exemplars $p_{m+1},\dots,p_{m'}$, 
keeping only the examples $p_1,\dots,p_m$.

\paragraph{Background\onedot} 
The exemplar management routines are designed with two objectives 
in mind: the initial exemplar set should approximate the class mean 
vector well, and it should be possible to remove exemplars at any 
time during the algorithm's runtime without violating this property.

The latter property is challenging because the actual class mean 
vector is not available to the algorithm anymore when the removal 
procedure is called. 
Therefore, we adopt a data-independent removal strategy,
removing elements in fixed order starting at the end,
and we make it the responsibility of the exemplar set 
construction routine to make sure that the desired 
approximation properties are fulfilled even after 
the removal procedure is called at later times.
The prioritized construction is the logical consequence 
of this condition: it ensures that the average feature 
vector over any subset of exemplars, starting at the 
first one, is a good approximation of the mean vector. 
The same prioritized construction is used 
in~\emph{herding}~\cite{Welling09} to create a 
representative set of samples from a distribution.
There it was also shown that the iterative selection 
requires fewer samples to achieve a high approximation 
quality than, \eg, random subsampling.
In contrast, other potential methods for exemplar 
selection, such as~\cite{elhamifar2013sparse,misra2014data},
were designed with other objectives and are not guaranteed 
to provide a good approximation quality for any number of
prototypes.

Overall, iCaRL's steps for exemplar selection and 
reduction fit exactly to the incremental learning setting: 
the selection step is required for each class only once, 
when it is first observed and its training data is available. 
At later times, only the reduction step is called, which does 
not need access to any earlier training data.

%%%%%%%%%%%%%%%%%%%%%%%%%%%%%%%%%%%%%%%%%%%%%%%%%%%%%%%%%%%%%%
%%%%%%%%%%%%%%%%%%%%%%%%%%%%%%%%%%%%%%%%%%%%%%%%%%%%%%%%%%%%%%
%%%%%%%%%%%\input{relatedwork.tex}
%%%%%%%%%%%%%%%%%%%%%%%%%%%%%%%%%%%%%%%%%%%%%%%%%%%%%%%%%%%%%%
%%%%%%%%%%%%%%%%%%%%%%%%%%%%%%%%%%%%%%%%%%%%%%%%%%%%%%%%%%%%%%
%

\section{Related work}\label{sec:relatedwork}
iCaRL builds on the insights of multiple earlier 
attempts to address class-incremental learning. 
In this section, we describe the most important ones, 
structuring them on the one hand into learning techniques 
with \emph{fixed data representations} and on the other
hand into techniques that also learn the data 
representation, both from the \emph{classical connectionists} 
era as well as recent \emph{deep learning} approaches. 

\paragraph{Learning with a fixed data representation.}
When the data representation is fixed, the main 
challenge for class-incremental learning is to 
design a classifier architecture that can accommodate new 
classes at any time during the training process without 
requiring access to all training data seen so far. 
The simplest such process of this type could be a 
\emph{(k-)nearest neighbor} classifier, but that would 
require storing all training data during the learning 
process and therefore does not qualify as a 
class-incremental procedure by our definition.

Mensink~\etal~\cite{Mensink2012} observed that the 
\emph{nearest class mean (NCM)} classifier has this 
property. 
NCM represents each class as a prototype vector that is the 
average feature vector of all examples observed for the 
class so far. 
This vector can be computed incrementally from a data 
stream, so there is no need to store all training examples. 
A new example is classified by assigning it the class 
label that has a prototype most similar to the example's 
feature vector, with respect to a metric that can also 
be learned from data.
Despite (or because of) its simplicity, NCM has been shown 
to work well and be more robust than standard parametric 
classifiers in an incremental 
learning setting~\cite{Mensink2012,mensink2013distance,ristin2014incremental}. 

NCM's main shortcoming is that it cannot easily be extended 
to the situation in which a nonlinear data representation should
be learned together with the classifiers, as this prevents 
the class mean vectors from being computable in an incremental
way. 
For iCaRL we adopt from NCM the idea of prototype-based classification. 
However, the prototypes we use are not the average features vectors 
over all examples but only over a specifically chosen subset, which 
allows us to keep a small memory footprint and perform all necessary 
updates with constant computational effort.

Alternative approaches fulfill the class-incremental learning 
criteria $i)$--$iii)$, that we introduced in Section~\ref{sec:intro},
only partially: 
Kuzborskij~\etal~\cite{kuzborskij2013n} showed that a loss of 
accuracy can be avoided when adding new classes to an existing 
linear multi-class classifier, as long as the classifiers can
be retrained from at least a small amount of data for all 
classes.
Chen~\etal~\cite{chen2013,chen2014} and Divvala~\etal~\cite{Divvala_2014_CVPR}
introduced systems that autonomously retrieve images from web resources 
and identifies relations between them, but they does not incrementally 
learn object classifiers. 
Royer and Lampert~\cite{royer-cvpr2015} adapt classifiers 
to a time-varying data stream
but their method cannot handle newly appearing classes, 
while Pentina~\etal~\cite{pentina-cvpr2015} show that 
learning multiple tasks sequentially can beneficial, but 
for choosing the order the data for all tasks has to be 
available at the same time.

Li and Wechsler~\cite{li2005open}, Scheirer~\etal~\cite{Scheirer_2013_TPAMI},
as well as Bendale and Boult~\cite{Bendale_2015_CVPR} aimed at the 
related but distinct problem of \emph{Open Set Recognition} in 
which test examples might come from other classes than the training 
examples seen so far. 
Polikar~\etal~\cite{MuhlbaierTP09,PolikarUUH01} introduced an 
ensemble based approach that can handle an increasing number of 
classes but needs training data for all classes to occur 
repeatedly.
Zero-shot learning, as proposed by Lampert~\etal~\cite{lampert-tpami2013}, 
can classify examples of previously unseen classes, 
but it does not include a training step for those.

\paragraph{Representation learning\onedot}
The recent success of (deep) neural networks can in large 
parts be attributed to their ability to learn not only 
classifiers but also suitable data representations~\cite{bengio2013representation, Li2016, misra2016cross,saxena2016convolutional}, 
at least in the standard batch setting.
First attempts to learn data representations in an incremental
fashion can already be found in the classic neural network
literature, \eg~\cite{ans1997avoiding,french1993,french1999catastrophic,robins95}. 
In particular, in the late 1980s \mbox{McCloskey}~\etal~\cite{mccloskey1989catastrophic} 
described the problem of \emph{catastrophic forgetting}, 
\ie the phenomenon that training a neural network with 
new data causes it to overwrite (and thereby forget) 
what it has learned on previous data.
However, these classical works were mainly in the context of 
connectionist memory networks, not classifiers, and the 
networks used were small and shallow by today's standards. 
Generally, the existing algorithms and architectural changes 
are unable to prevent catastrophic forgetting, see, for example, Moe-Helgesen~\etal's survey~\cite{moe2005catastophic} 
for classical and Goodfellow~\etal's~\cite{goodfellow14empirical} 
for modern architectures, except in specific settings, such as Kirkpatrick~\etal's~\cite{kirkpatrick2017overcoming}. 

A major achievement of the early connectionist works, however, is 
that they identified the two main strategies of how catastrophic 
forgetting can be addressed: 1) by \emph{freezing} parts of the 
network weights while at the same time \emph{growing} 
the network in order to preserve the 
ability to learn, 2) by \emph{rehearsal}, \ie continuously stimulating 
the network not only with the most recent, but also with earlier data.

Recent works on incremental learning of neural networks have mainly 
followed the freeze/grow strategy, which however requires allocating 
more and more resources to the network over time and therefore 
violates principle $iii)$ 
of our definition of class-incremental learning.
For example, Xiao~\etal~\cite{xiao2014error} learn a tree-structured 
model that grows incrementally as more classes are observed. 
In the context of multi-task reinforcement learning, 
Rusu~\etal~\cite{rusu2016progressive} propose growing the 
networks by extending all layer horizontally.

For iCaRL, we adopt the principle of \emph{rehearsal}: to update 
the model parameters for learning a representation, we use not only 
the training data for the currently available classes, but also the 
exemplars from earlier classes, which are available anyway as they 
are required for the prototype-based classification rule.
Additionally, iCaRL also uses \emph{distillation} to prevent that 
information in the network deteriorates too much over time.
while Hinton~\etal~\cite{hinton2015distilling} originally proposed 
distillation to transfer information between different 
neural networks, in iCaRL, we use it 
within a single network between different time points. 
The same principle was recently proposed by Li and Hoiem~\cite{Li2016}
under the name of \emph{Learning without Forgetting (LwF)} to 
incrementally train a single network for learning multiple 
tasks, \eg multiple object recognition datasets. 
The main difference to the class-incremental multi-class situation 
lies in the prediction step: a multi-class learner has to pick 
one classifier that predicts correctly any of the observed classes. 
A multi-task (multi-dataset) leaner can make use of multiple classifiers,
each being evaluated only on the data from its own dataset. 

%%%%%%%%%%%%%%%%%%%%%%%%%%%%%%%%%%%%%%%%%%%%%%%%%%%%%%%%%%%%%%
%%%%%%%%%%%%%%%%%%%%%%%%%%%%%%%%%%%%%%%%%%%%%%%%%%%%%%%%%%%%%%
%%%%%%%%%%%\input{experiments.tex}
%%%%%%%%%%%%%%%%%%%%%%%%%%%%%%%%%%%%%%%%%%%%%%%%%%%%%%%%%%%%%%
%%%%%%%%%%%%%%%%%%%%%%%%%%%%%%%%%%%%%%%%%%%%%%%%%%%%%%%%%%%%%%

\begin{figure*}[t]\centering
\subfloat[Multi-class accuracy (averages and standard deviations over 10 repeats) on 
iCIFAR-100 with 2 (top left), \quad 5 (top middle), 10 (top right), 20 (bottom left) or 
50 (bottom right) classes per batch.] 
{
\begin{tabular}{c}\includegraphics[height=.25\textwidth]{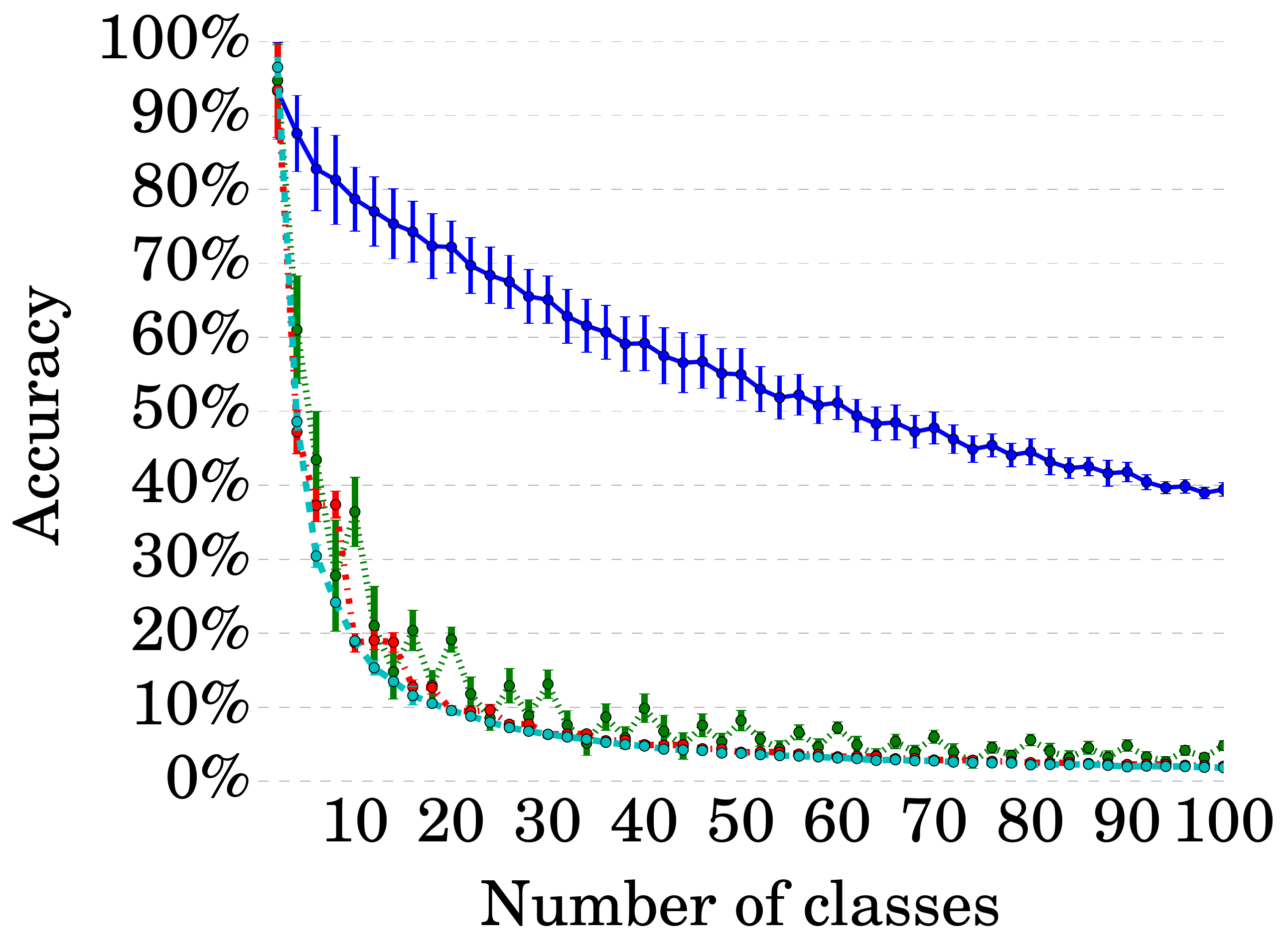}
\includegraphics[height=.25\textwidth]{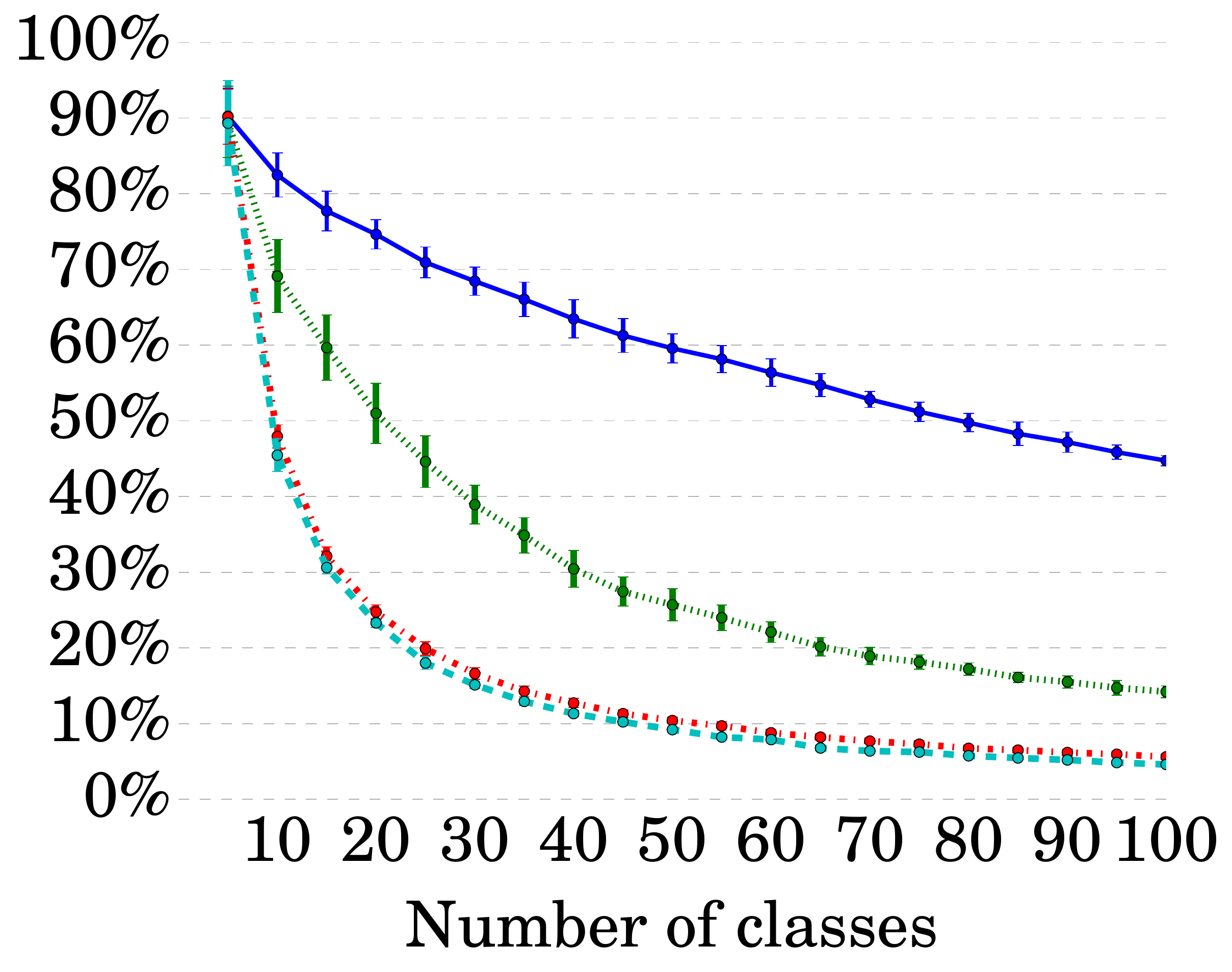}
\includegraphics[height=.25\textwidth]{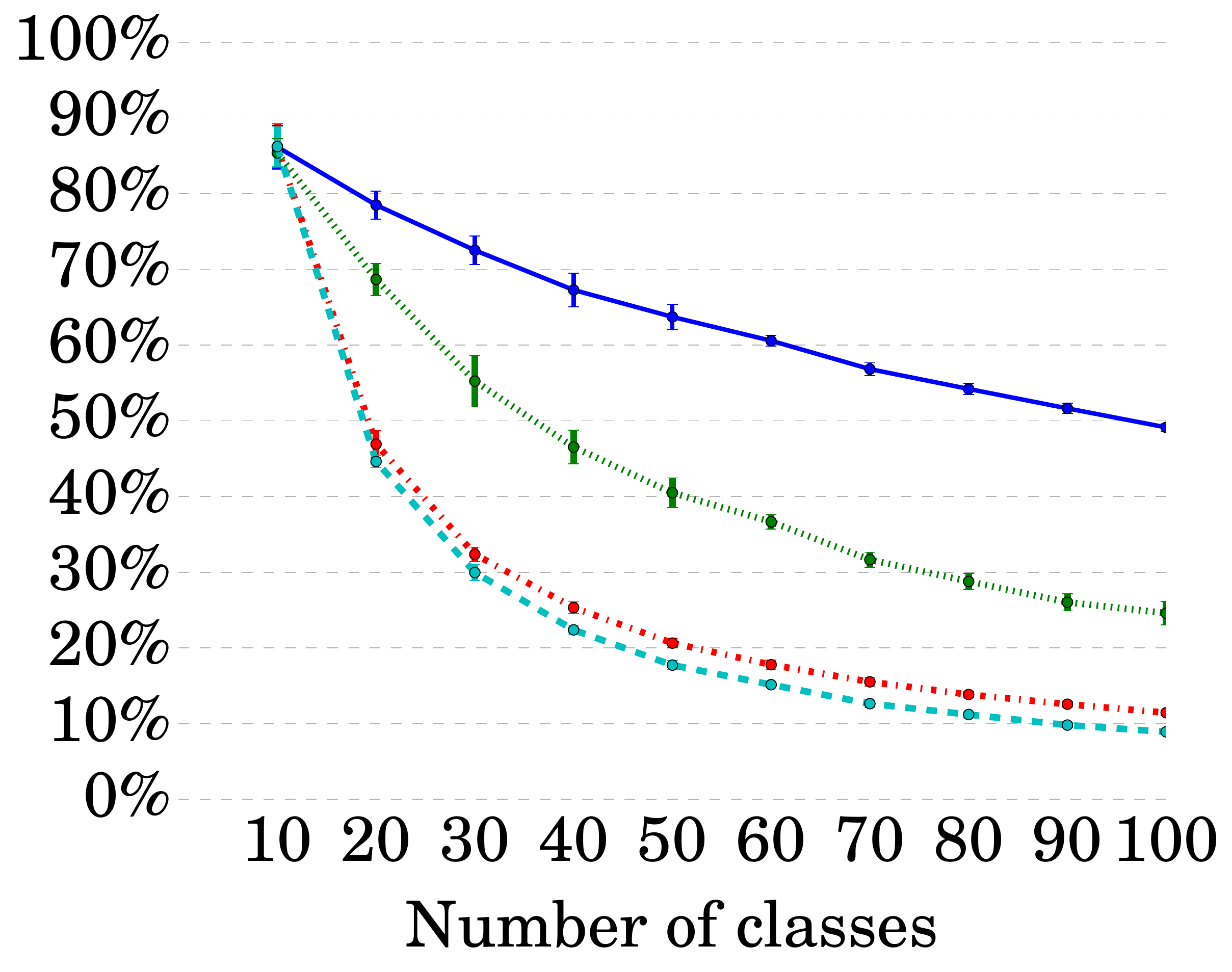}
\\
\includegraphics[height=.25\textwidth]{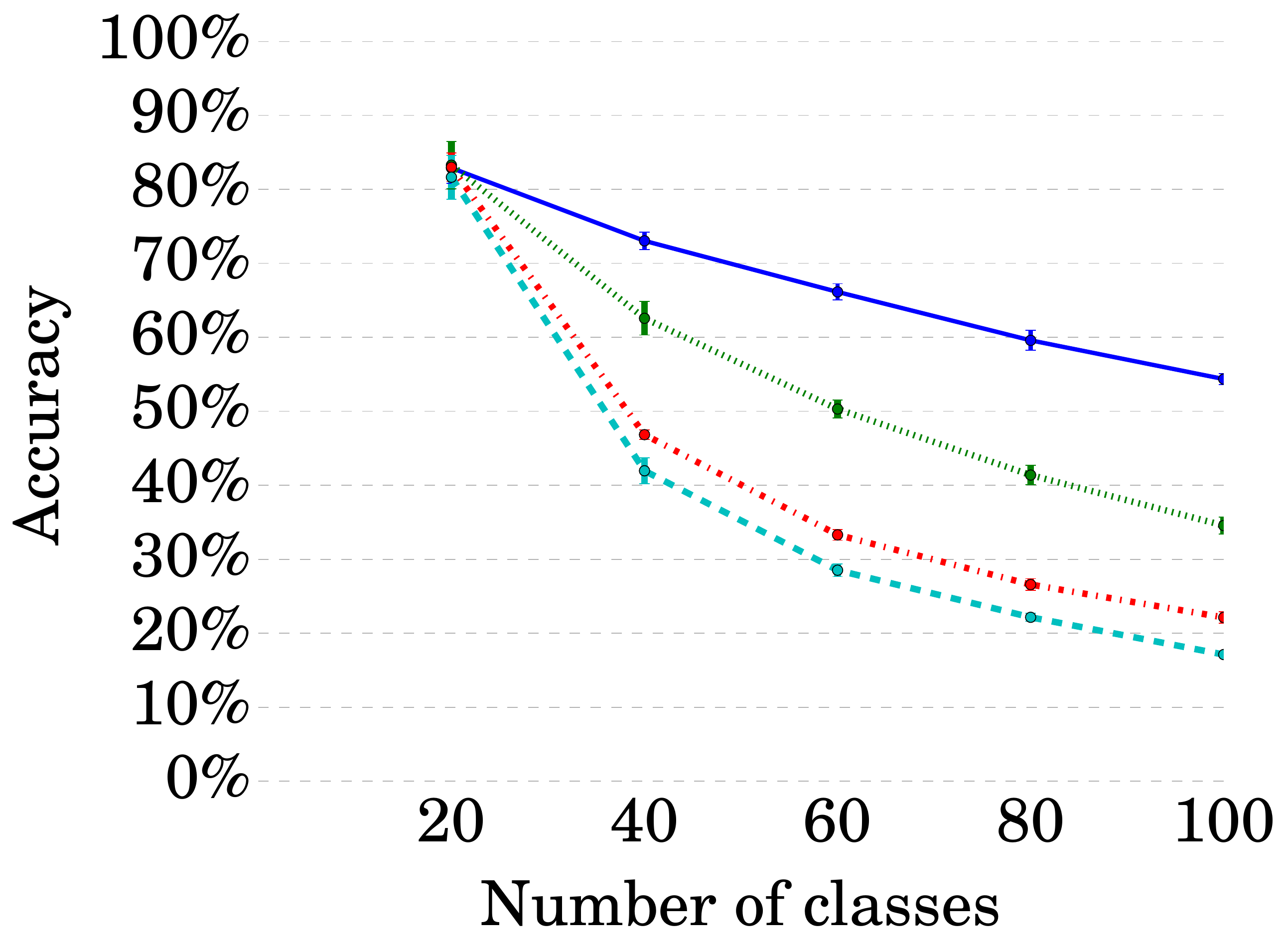}
\includegraphics[height=.25\textwidth]{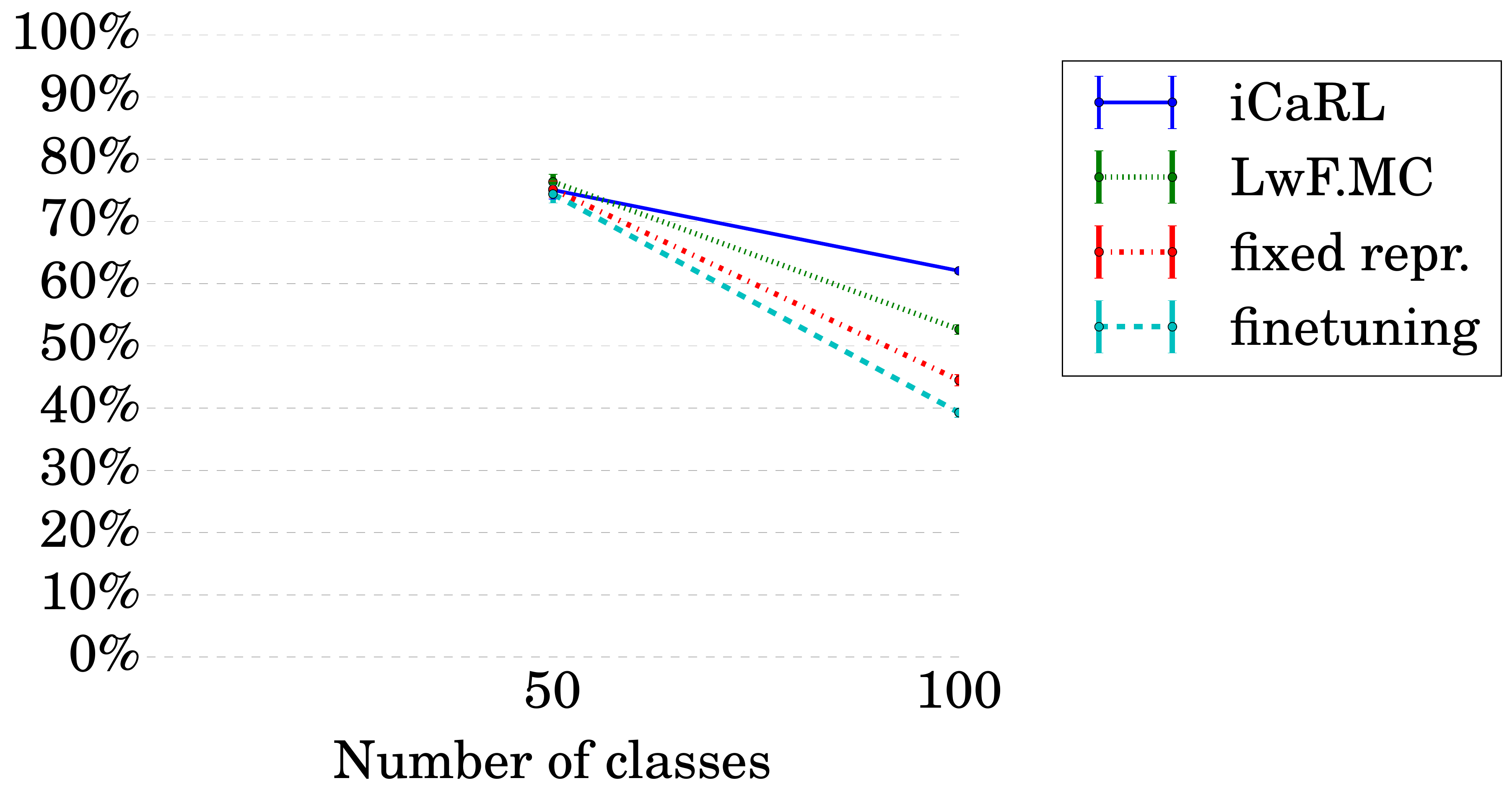}
\end{tabular}
}\quad
\subfloat[Top-5 accuracy on iILSVRC-small (top) and iILSVRC-full (bottom). ]
{
\includegraphics[height=.28\textwidth]{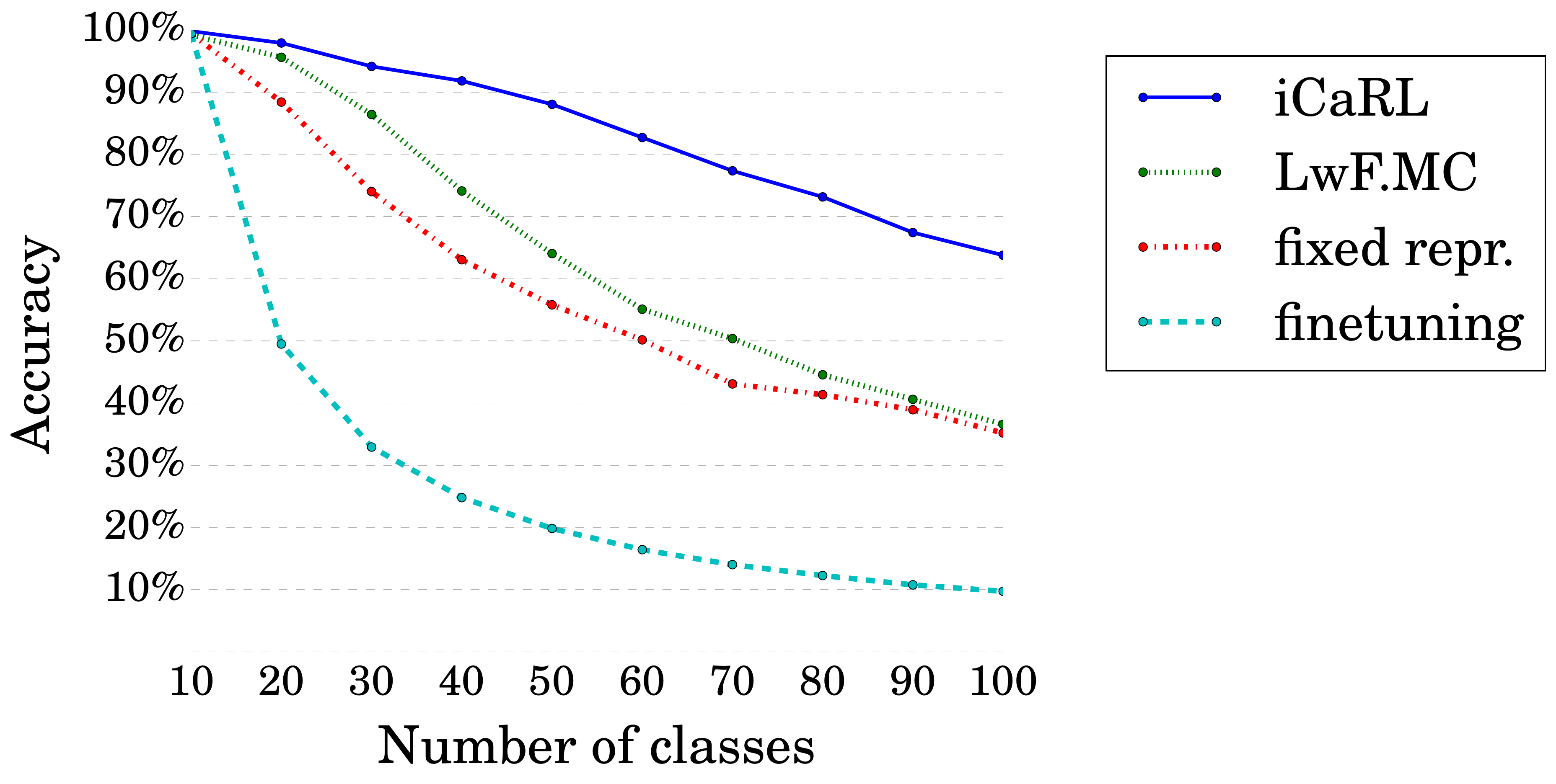}\quad
\includegraphics[height=.28\textwidth]{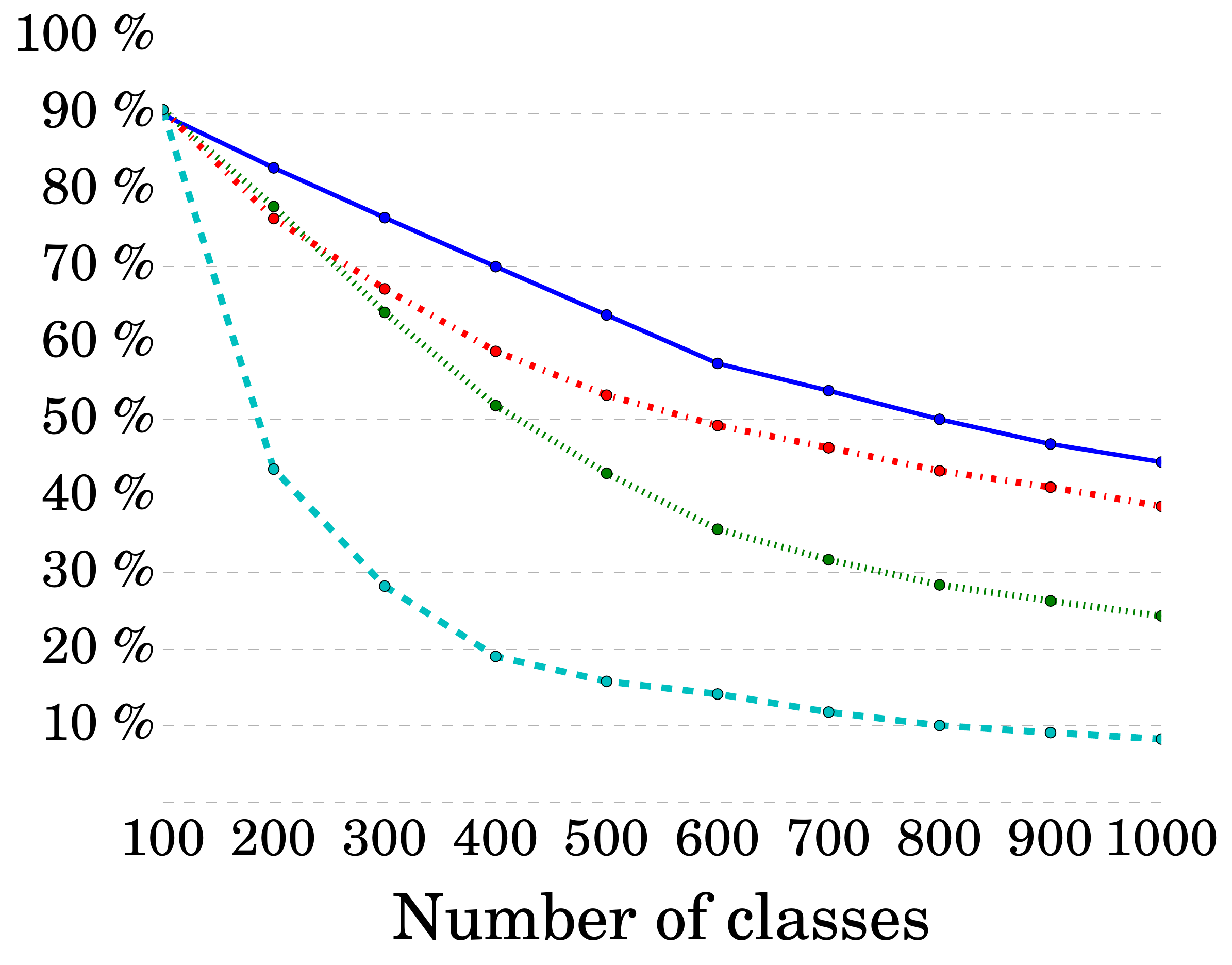}
}
\caption{Experimental results of class-incremental training on iCIFAR-100 and iILSVRC: 
reported are multi-class accuracies across all classes observed up to a certain 
time point. 
iCaRL clearly outperforms the other methods in this setting. 
Fixing the data representation after having trained on the first batch (\emph{fixed repr.})
performs worse than distillation-based LwF.MC, except for \emph{iILSVRC-full}.
Finetuning the network without preventing catastrophic forgetting (\emph{finetuning}) 
achieves the worst results. 
For comparison, the same network trained with all data available achieves 68.6\% multi-class accuracy.
}\label{fig:results}
\end{figure*}

\begin{figure*}[t]
\centering
\subfloat[iCaRL]{\includegraphics[height=.43\textwidth]{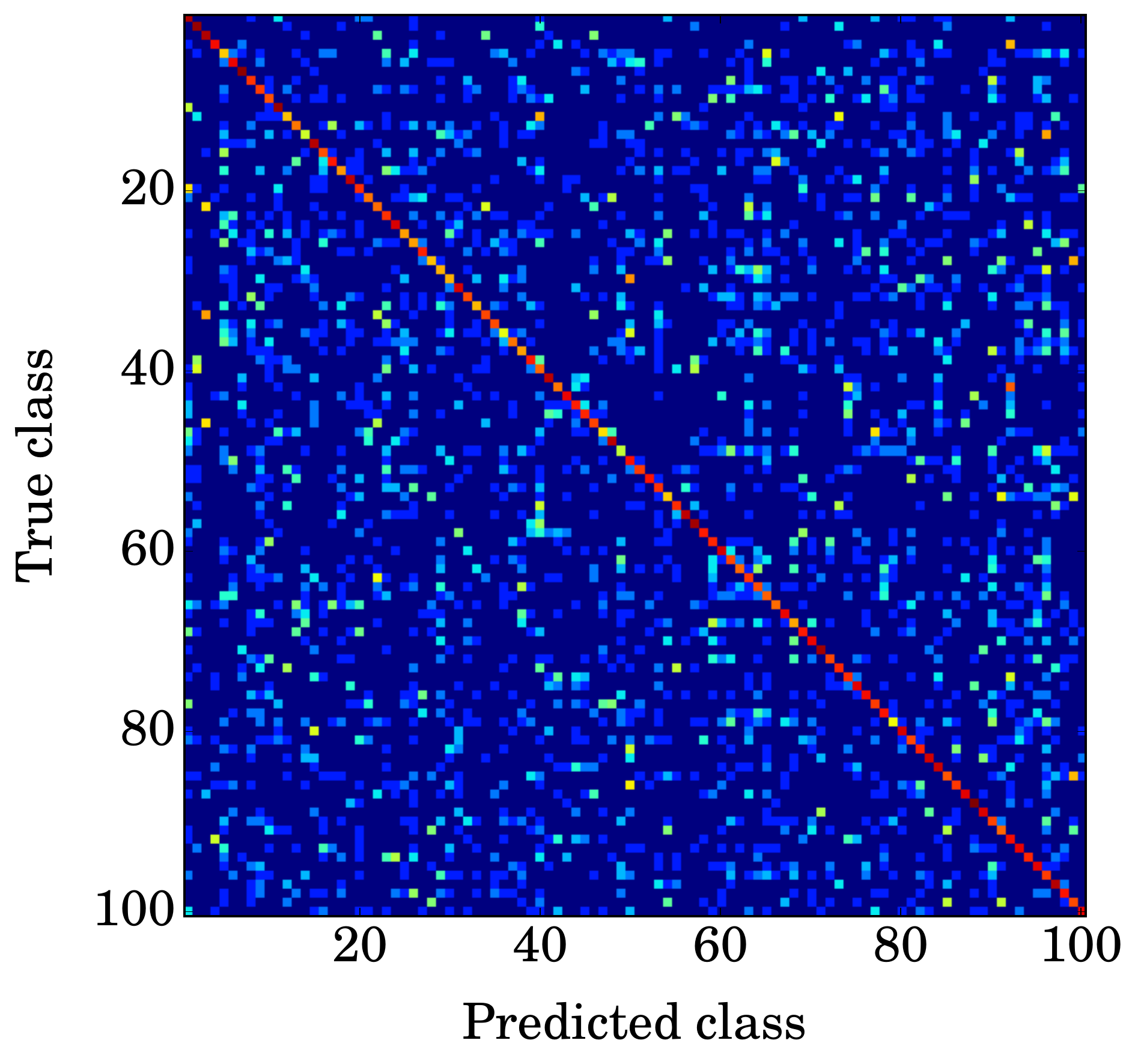}}\qquad 
\subfloat[LwF.MC]{\includegraphics[height=.43\textwidth]{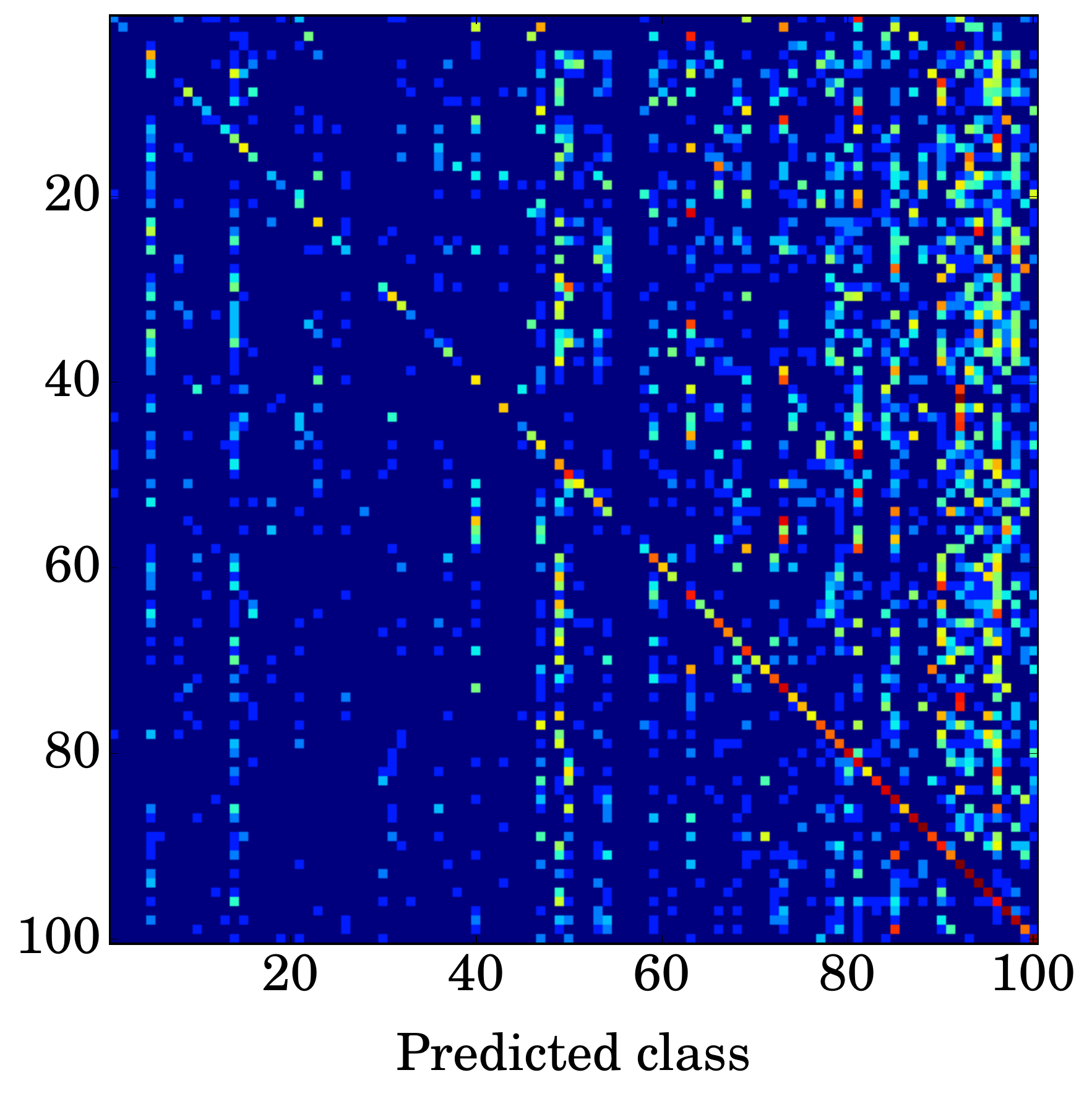}}
\\
\subfloat[fixed representation]{\includegraphics[height=.43\textwidth]{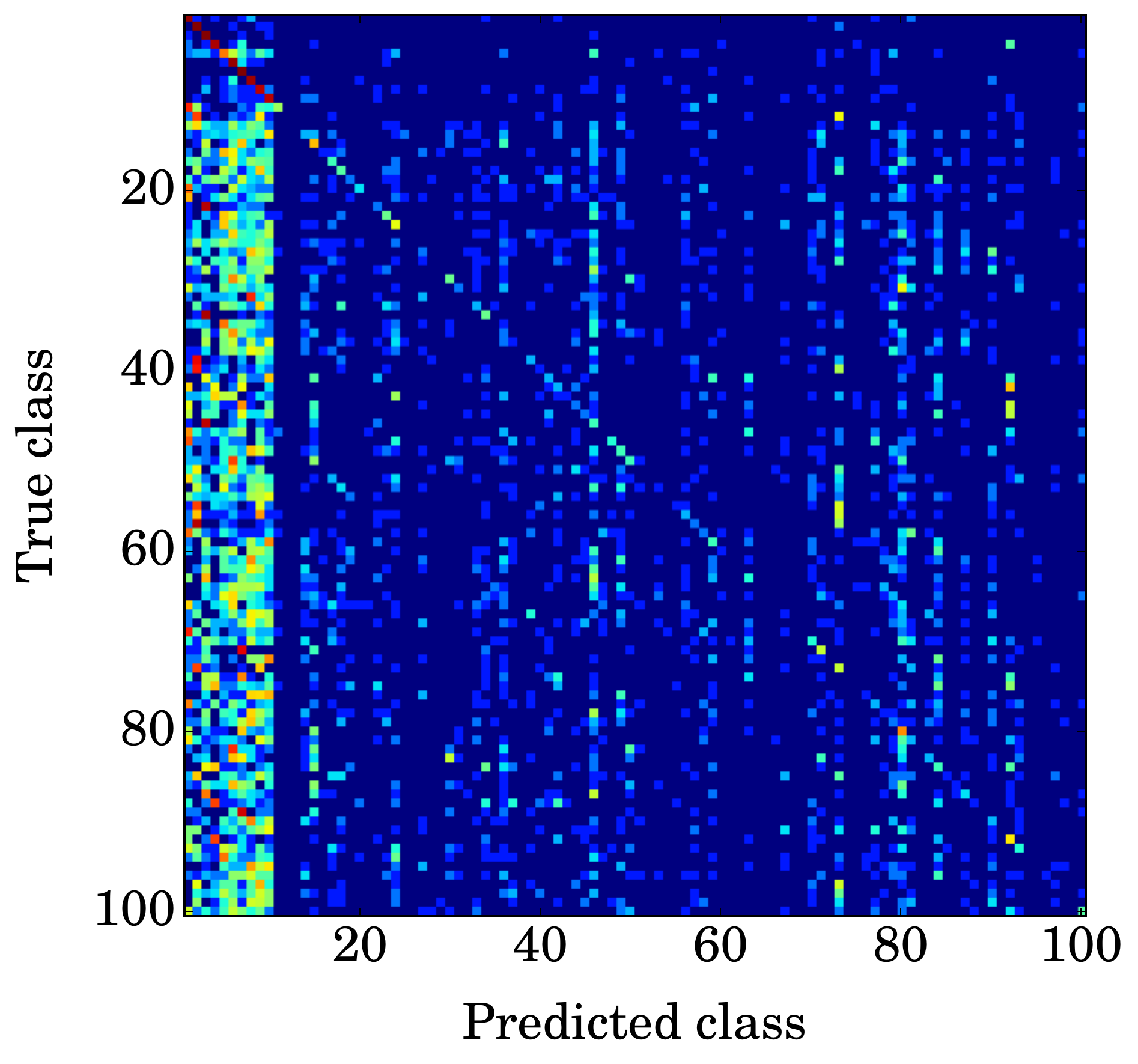}}\qquad
\subfloat[finetuning]{\includegraphics[height=.43\textwidth]{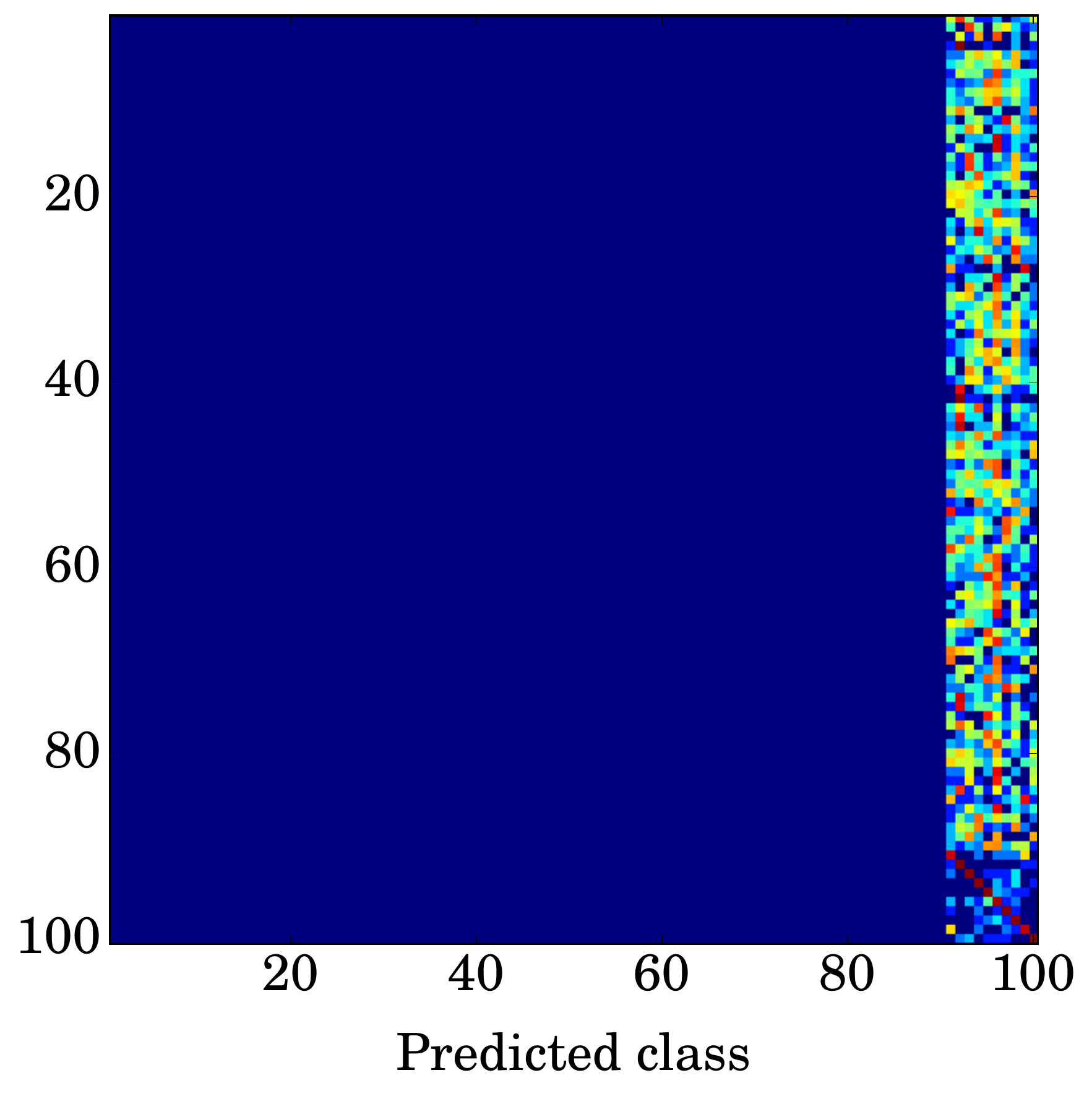}}
\caption{Confusion matrices of different method on iCIFAR-100 (with entries transformed by $\log(1+x)$ 
for better visibility). 
iCaRL's predictions are distributed close to uniformly over all classes, 
whereas LwF.MC tends to predict classes from recent batches more 
frequently. 
The classifier with fixed representation has a bias towards classes from 
the first batch, while the network trained by finetuning predicts exclusively 
classes labels from the last batch.
}\label{fig:confusionmatrices}
\end{figure*}

\section{Experiments}\label{sec:experiments}
In this section we propose a protocol for evaluating 
incremental learning methods and compare iCaRL's classification 
accuracy to that of alternative methods (Section~\ref{subsec:results}). 
We also report on further experiments that shed light on iCaRL's 
working mechanisms by isolating the effect of individual components 
(Section~\ref{subsec:study}).

\paragraph{Benchmark protocol\onedot}
So far, no agreed upon benchmark protocol for evaluation 
class-incremental learning methods exist. Therefore, we 
propose the following evaluation procedure: 
for a given multi-class classification dataset, the classes are 
arranged in a fixed random order. Each method is then trained in 
a class-incremental way on the available training data. 
After each batch of classes, the resulting classifier is evaluated 
on the test part data of the dataset, considering only those classes 
that have already been trained. Note that, even though the test data 
is used more than once, no overfitting can occur, as the testing 
results are not revealed to the algorithms. 
The result of the evaluation are curves of the classification 
accuracies after each batch of classes. 
If a single number is preferable, we report the average of 
these accuracies, called \emph{average incremental accuracy}.

For the task of image classification we introduce two 
instantiations of the above protocol. 
1) \emph{iCIFAR-100 benchmark}: we use the CIFAR-100~\cite{krizhevsky2009learning} 
data and train all 100 classes in batches of 2, 5, 10, 20 or 50 classes 
at a time. The evaluation measure is the standard multi-class accuracy 
on the test set. 
As the dataset is of manageable size, we run this benchmark ten times 
with different class orders and reports averages and standard deviations 
of the results.
2) \emph{iILSVRC benchmark}: we use the ImageNet ILSVRC\,2012~\cite{ILSVRC15} 
dataset in two settings: using only a subset of 100 classes, which are trained 
in batches of 10 (\emph{iILSVRC-small}) or using all 1000 
classes, processed in batches of 100 (\emph{iILSVRC-full}). 
The evaluation measure is the \emph{top-5} accuracy on 
the \emph{val} part of the dataset.

\paragraph{iCaRL implementation.}
For iCIFAR-100 we rely on the \emph{theano} package\footnote{\url{http://deeplearning.net/software/theano/}} and train a 
32-layers ResNet~\cite{He2015},
allowing iCaRL to store up to $K=2000$ exemplars. 
Each training step consists of 70 epochs. The learning rate starts 
at 2.0 and is divided by 5 after 49 and 63 epochs (7/10 and 9/10 of all epochs).
For iILSVRC the maximal number of exemplars is $K\!=20000$ and 
we use the \emph{tensorflow} framework\footnote{\url{https://www.tensorflow.org/}} 
to train an 18-layers ResNet~\cite{He2015} for 60 epochs per class batch.
The learning rate starts at 2.0 and is divided by 5 after 20, 30, 40 and 50 epochs
(1/3, 1/2, 2/3 and 5/6 of all epochs).
For all methods we train the network using standard backpropagation 
with minibatches of size 128 and a weight decay parameter of $0.00001$.
Note that the learning rates might appear large, but for our purpose 
they worked well, likely because we use binary cross-entropy in 
the network layer. 
Smaller rates might be required for a multi-class softmax layer. 
Our source code and further data are available at \url{http://www.github.com/srebuffi/iCaRL}.

\subsection{Results}\label{subsec:results}
Our main set of experiments studies the classification 
accuracy of different methods under class-incremental 
conditions. Besides iCaRL we implemented and tested 
three alternative class-incremental methods. \emph{Finetuning} learns 
an ordinary multi-class network without taking any measures to 
prevent catastrophic forgetting. It can also be interpreted as 
learning a multi-class classifier for new incoming classes by 
finetuning the previously learned multiclass classification network. 
\emph{Fixed representation} also learns a multi-class classification 
network, but in a way that prevents catastrophic forgetting. It freezes 
the feature representation after the first batch of classes has been 
processed and the weights of the classification layer after the 
corresponding classes have been processed. 
For subsequent batches of classes, only the weights vectors of 
new classes are trained. 
Finally, we also compare to a network classifier that attempts at 
preventing catastrophic forgetting by using the distillation loss 
during learning, like iCaRL does, but that does not use an exemplar 
set. 
For classification, it uses the network output values themselves. 
This is essentially the \emph{Learning without Forgetting} approach,
but applied to multi-class classification we, so denote it by LwF.MC.
Figure~\ref{fig:results} shows the results.
One can see that iCaRL clearly outperforms the other methods, 
and the more so the more incremental the setting is (\ie 
the fewer classes can be processed at the same time). 
Among the other methods, \emph{distillation}-based network training 
(LwF.MC) is always second best, except for \emph{iILSVRC-full},
where it is better to fix the representation after the first 
batch of 100 classes. 
\emph{Finetuning} always achieves the worst results, confirming 
that catastrophic forgetting is indeed a major problem for in 
class-incremental learning. 

Figure~\ref{fig:confusionmatrices} provides further insight into 
the behavior of the different methods. Is shows the confusion 
matrices of the 100-class classifier on iCIFAR-100 
after training using batches of 10 classes at a time (larger versions 
can be found in the appendix). 
One can see very characteristic patterns: iCaRL's confusion matrix 
looks homogeneous over all classes, both in terms of the diagonal 
entries (\ie correct predictions) as well as off-diagonal entries 
(\ie mistakes). 
This shows that iCaRL has no intrinsic bias towards or against 
classes that it encounters early or late during learning. In 
particular, it does not suffer from catastrophic forgetting.

In contrast to this, the confusion matrices for the other classes 
show inhomogeneous patterns: \emph{distillation}-based training 
(LwF.MC) has many more non-zero entries towards the right, \ie for 
recently learned classes. 
Even more extreme is the effect for \emph{finetuning}, where all 
predicted class labels come from the last batch of classes that 
the network has been trained with. The finetuned network 
simply \emph{forgot} that earlier classes even exist. 
The \emph{fixed representation} shows the opposite pattern: it 
prefers to output classes from the first batch of classes it was 
trained on (which were used to obtained the data representation). 
Confusion matrices for iILSVRC show the same patterns, they can be found in the appendix.

\begin{table*}
\caption{Average multi-class accuracy on iCIFAR-100 for different modifications of iCaRL.}\label{tab:details}
\centering
\subfloat[Switching off different components of iCaRL 
(\emph{hybrid1}, \emph{hybrid2}, \emph{hybrid3}, see text for details)
leads to results mostly inbetween iCaRL and LwF.MC, showing that all 
of iCaRL's new components contribute to its performance. %See the main text for a further analysis.
]{\small
\begin{tabular}{c|c|c|c|c|c}
batch size & iCaRL   &  \emph{hybrid1} & \emph{hybrid2} & \emph{hybrid3} & LwF.MC \\\hline
2 classes &  57.0  &  36.6 &  57.6  & 57.0 &  11.7
\\
5 classes&  61.2   &  50.9 &  57.9  & 56.7 & 32.6
\\
10 classes& 64.1   &  59.3 & 59.9   & 58.1 & 44.4
\\
20 classes& 67.2    &  65.6 &  63.2  & 60.5 & 54.4
\\
50 classes& 68.6    &  68.2 &  65.3  & 61.5 & 64.5
\end{tabular}\label{tab:details1}}
\quad\ \ \ 
\subfloat[Replacing iCaRL's mean-of-exemplars by a nearest-class-mean 
classifier (NCM) has only a small positive effect on the classification 
accuracy, showing that iCaRL's strategy for selecting exemplars is 
effective.]{\small
\hspace*{2.cm}\begin{tabular}{c|c|c}
batch size & iCaRL  & NCM   \\\hline
2 classes  &  57.0  &  59.3
\\
5 classes &  61.2  &  62.1
\\
10 classes & 64.1  &  64.5
\\
20 classes& 67.2  &  67.5
\\
50 classes& 68.6  &  68.7
\end{tabular}\hspace*{2.cm}\label{tab:details2}}
\end{table*}

\subsection{Differential Analysis}\label{subsec:study}

To provide further insight into the working mechanism 
of iCaRL, we performed additional experiments on 
iCIFAR-100, in which we isolate individual aspects of 
the methods. 

First, we analyze why exactly iCaRL improves over plain 
finetuning-based training, from which it differs in 
three aspects: by the use of the mean-of-exemplars 
classification rule, by the use of exemplars during 
the representation learning, and by the use of the
distillation loss.
We therefore created three hybrid setups: the first 
\emph{(hybrid1)} learns a representation in the same way 
as iCaRL, but uses the network's outputs directly for 
classification, not the mean-of-exemplar classifier.
The second \emph{(hybrid2)} uses the exemplars for 
classification, but does not use the distillation 
loss during training.
The third \emph{(hybrid3)} uses neither the distillation 
loss nor exemplars for classification, but it makes use 
of the exemplars during representation learning.
For comparison, we also include LwF.MC again, which 
uses distillation, but no exemplars at all. 

Table~\ref{tab:details1} summarizes the results as the 
average of the classification accuracies over all 
steps of the incremental training.
One can see that the hybrid setups mostly achieve 
results in between iCaRL and LwF.MC, showing that 
indeed all of iCaRL's new components 
contribute substantially to its good performance. 
In particular, the comparison of iCaRL with \emph{hybrid1} shows 
that the mean-of-exemplar classifiers is particularly advantageous
for smaller batch sizes, \ie when more updates of the 
representation are performed.
Comparing iCaRL and \emph{hybrid2} one sees that for very small 
class batch sizes, distillation can even hurt classification 
accuracy compared to just using prototypes. 
For larger batch sizes and fewer updates, the use of the 
distillation loss is clearly advantageous. 
Finally, comparing the result of \emph{hybrid3} with LwF.MC 
clearly shows the effectiveness of exemplars in preventing 
catastrophic forgetting.

In a second set of experiments we study how much accuracy 
is lost by using the means-of-exemplars as classification 
prototypes instead of the nearest-class-mean \emph{(NCM)} rule.
For the latter, we use the unmodified iCaRL to learn a 
representation, but we classify images with NCM, where 
the class-means are recomputed after each representation 
update using the current feature extractor.
Note that this requires storing all training data,
so it would not qualify as a class-incremental method.
The results in Table~\ref{tab:details2} show only 
minor differences between iCaRL and NCM, confirming 
that iCaRL reliably identifies representative exemplars. 

\begin{figure}[t]\centering
\includegraphics[width=\columnwidth]{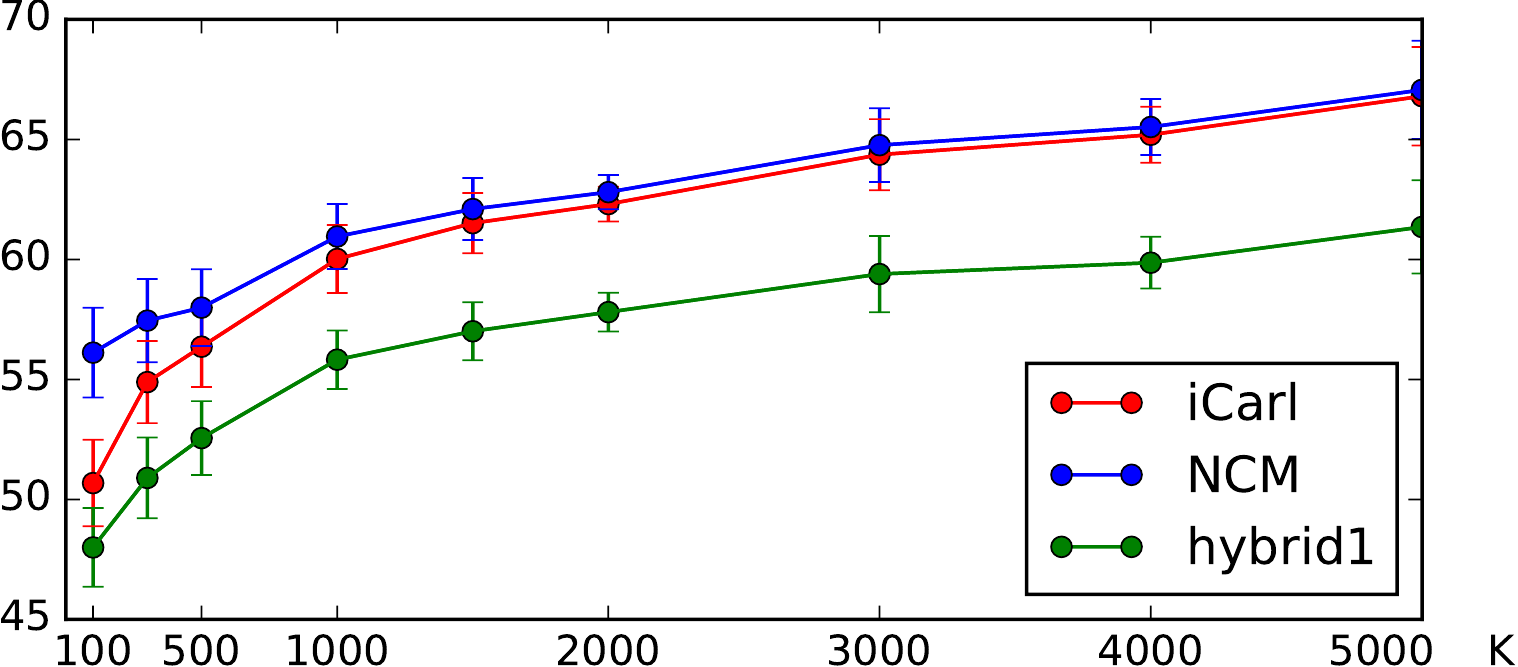}
\caption{Average incremental accuracy on iCIFAR-100 with 10 classes per batch 
for different memory budgets $K$.}\label{fig:K}
\end{figure}

Figure~\ref{fig:K} illustrates the effect of different memory budgets, 
comparing iCaRL with the \emph{hybrid1} classifier of Table~\ref{tab:details1} 
and the NCM classifier of Table~\ref{tab:details2}. 
Both use the same data representation as iCaRL 
but differ in their classification rules. 
All method benefit from a larger memory budget, showing that 
iCaRL's representation learning step indeed benefits from more 
prototypes.
Given enough prototypes (here at least 1000), iCaRL's 
mean-of-exemplars classifier performs similarly to the NCM classifier,
while classifying by the network outputs is not competitive.
%

%%%%%%%%%%%%%%%%%%%%%%%%%%%%%%%%%%%%%%%%%%%%%%%%%%%%%%%%%%%%%%
%%%%%%%%%%%%%%%%%%%%%%%%%%%%%%%%%%%%%%%%%%%%%%%%%%%%%%%%%%%%%%
%%%%%%%%%%%\input{conclusion.tex}
%%%%%%%%%%%%%%%%%%%%%%%%%%%%%%%%%%%%%%%%%%%%%%%%%%%%%%%%%%%%%%
%%%%%%%%%%%%%%%%%%%%%%%%%%%%%%%%%%%%%%%%%%%%%%%%%%%%%%%%%%%%%%
\section{Conclusion}\label{sec:conclusion}
We introduced iCaRL, a strategy for class-incremental learning 
that learns classifiers and a feature representation simultaneously.
iCaRL's three main components are: 1) a \emph{nearest-mean-of-exemplars}
classifier that is robust against changes in the data representation
while needing to store only a small number of exemplars per class, 
2) a \emph{herding}-based step for prioritized exemplar selection, 
and 3) a representation learning step that uses the exemplars in
combination with \emph{distillation} to avoid catastrophic forgetting.
Experiments on CIFAR-100 and ImageNet ILSVRC~2012 data show that iCaRL 
is able to learn incrementally over a long period of time where other 
methods fail quickly.

The main reason for iCaRL's strong classification results are its 
use of exemplar images.
While it is intuitive that being able to rely on stored exemplars 
in addition to the network parameters could be beneficial, we 
nevertheless find it an important observation how pronounced this 
effect is in the class-incremental setting.
We therefore hypothesize that also other architectures should be
able to benefit from using a combination of network parameters 
and exemplars, especially given the fact that many thousands of 
images can be stored (in compressed form) with memory requirements 
comparable to the sizes of current deep networks. 

Despite the promising results, class-incremental classification is 
far from solved. In particular, iCaRL's performance is still lower 
than what systems achieve when trained in a batch setting, \ie with 
all training examples of all classes available at the same time. 
In future work we plan to analyze the reasons for this in more 
detail with the goal of closing the remaining performance gap.
We also plan to study related scenarios in which the classifier 
cannot store any of the training data in raw form, \eg for privacy 
reasons.
A possible direction for this would be to encode feature 
characteristics of earlier tasks implicitly by a autoencoder, 
as recently proposed by Rannen Triki~\etal~\cite{rannen2017b}.

%%%%%%%%%%%%%%%%%%%%%%%%%%%%%%%%%%%%%%%%%%%%%%%%%%%%%%%%%%%%%%
%%%%%%%%%%%%%%%%%%%%%%%%%%%%%%%%%%%%%%%%%%%%%%%%%%%%%%%%%%%%%%
%%%%%%%%%%%\input{acknowledgement.tex}
%%%%%%%%%%%%%%%%%%%%%%%%%%%%%%%%%%%%%%%%%%%%%%%%%%%%%%%%%%%%%%
%%%%%%%%%%%%%%%%%%%%%%%%%%%%%%%%%%%%%%%%%%%%%%%%%%%%%%%%%%%%%%
\paragraph{Acknowledgments.} 
This work was in parts funded by the European Research Council 
under the European Union's Seventh Framework Programme (FP7/2007-2013)/ERC 
grant agreement no 308036: "Life-long learning of visual scene understanding" 
(L3ViSU). 
The Tesla K40 cards used for this research were donated by the NVIDIA Corporation.

{\small
\bibliographystyle{ieee}
\bibliography{icarl-arxiv}

\begin{thebibliography}{10}\itemsep=-1pt

\bibitem{ans1997avoiding}
B.~Ans and S.~Rousset.
\newblock Avoiding catastrophic forgetting by coupling two reverberating neural
  networks.
\newblock {\em Comptes Rendus de l'Acad{\'e}mie des Sciences}, 320(12), 1997.

\bibitem{Bendale_2015_CVPR}
A.~Bendale and T.~Boult.
\newblock Towards open world recognition.
\newblock In {\em Conference on Computer Vision and Pattern Recognition
  (CVPR)}, 2015.

\bibitem{bengio2013representation}
Y.~Bengio, A.~Courville, and P.~Vincent.
\newblock Representation learning: A review and new perspectives.
\newblock {\em IEEE Transactions on Pattern Analysis and Machine Intelligence
  (T-PAMI)}, 35(8), 2013.

\bibitem{chen2013}
X.~Chen, A.~Shrivastava, and A.~Gupta.
\newblock {NEIL}: {E}xtracting visual knowledge from web data.
\newblock In {\em International Conference on Computer Vision (ICCV)}, 2013.

\bibitem{chen2014}
X.~Chen, A.~Shrivastava, and A.~Gupta.
\newblock Enriching visual knowledge bases via object discovery and
  segmentation.
\newblock In {\em Conference on Computer Vision and Pattern Recognition
  (CVPR)}, 2014.

\bibitem{Divvala_2014_CVPR}
S.~K. Divvala, A.~Farhadi, and C.~Guestrin.
\newblock Learning everything about anything: Webly-supervised visual concept
  learning.
\newblock In {\em Conference on Computer Vision and Pattern Recognition
  (CVPR)}, 2014.

\bibitem{elhamifar2013sparse}
E.~Elhamifar and R.~Vidal.
\newblock Sparse subspace clustering: Algorithm, theory, and applications.
\newblock {\em IEEE Transactions on Pattern Analysis and Machine Intelligence
  (T-PAMI)}, 35(11):2765--2781, 2013.

\bibitem{french1993}
R.~M. French.
\newblock Catastrophic interference in connectionist networks: Can it be
  predicted, can it be prevented?
\newblock In {\em Conference on Neural Information Processing Systems (NIPS)},
  1993.

\bibitem{french1999catastrophic}
R.~M. French.
\newblock Catastrophic forgetting in connectionist networks.
\newblock {\em Trends in cognitive sciences}, 3(4), 1999.

\bibitem{goodfellow14empirical}
I.~J. Goodfellow, M.~Mirza, D.~Xiao, A.~Courville, and Y.~Bengio.
\newblock An empirical investigation of catastrophic forgeting in
  gradient-based neural networks.
\newblock In {\em International Conference on Learning Representations (ICLR)},
  2014.

\bibitem{He2015}
K.~He, X.~Zhang, S.~Ren, and J.~Sun.
\newblock Deep residual learning for image recognition.
\newblock {\em arXiv preprint arXiv:1512.03385}, 2015.

\bibitem{hinton2015distilling}
G.~Hinton, O.~Vinyals, and J.~Dean.
\newblock Distilling the knowledge in a neural network.
\newblock In {\em NIPS Workshop on Deep Learning}, 2014.

\bibitem{IoffeS15}
S.~Ioffe and C.~Szegedy.
\newblock Batch normalization: Accelerating deep network training by reducing
  internal covariate shift.
\newblock In {\em International Conference on Machine Learing (ICML)}, 2015.

\bibitem{KingmaB14}
D.~P. Kingma and J.~Ba.
\newblock Adam: {A} method for stochastic optimization.
\newblock In {\em International Conference on Learning Representations (ICLR)},
  2015.

\bibitem{kirkpatrick2017overcoming}
J.~Kirkpatrick, R.~Pascanu, N.~Rabinowitz, J.~Veness, G.~Desjardins, A.~A.
  Rusu, K.~Milan, J.~Quan, T.~Ramalho, A.~Grabska-Barwinska, et~al.
\newblock Overcoming catastrophic forgetting in neural networks.
\newblock {\em Proceedings of the National Academy of Sciences (PNAS)}, 2017.

\bibitem{krizhevsky2009learning}
A.~Krizhevsky.
\newblock Learning multiple layers of features from tiny images.
\newblock Technical report, University of Toronto, 2009.

\bibitem{kuzborskij2013n}
I.~Kuzborskij, F.~Orabona, and B.~Caputo.
\newblock From $n$ to $n+1$: Multiclass transfer incremental learning.
\newblock In {\em Conference on Computer Vision and Pattern Recognition
  (CVPR)}, 2013.

\bibitem{lampert-tpami2013}
C.~H. Lampert, H.~Nickisch, and S.~Harmeling.
\newblock Attribute-based classification for zero-shot visual object
  categorization.
\newblock {\em IEEE Transactions on Pattern Analysis and Machine Intelligence
  (T-PAMI)}, 2013.

\bibitem{LeCun1998}
Y.~LeCun, L.~Bottou, Y.~Bengio, and P.~Haffner.
\newblock Gradient-based learning applied to document recognition.
\newblock {\em Proceedings of the IEEE}, 86(11), 1998.

\bibitem{li2005open}
F.~Li and H.~Wechsler.
\newblock Open set face recognition using transduction.
\newblock {\em IEEE Transactions on Pattern Analysis and Machine Intelligence
  (T-PAMI)}, 27(11), 2005.

\bibitem{Li2016}
Z.~Li and D.~Hoiem.
\newblock Learning without forgetting.
\newblock In {\em European Conference on Computer Vision (ECCV)}, 2016.

\bibitem{mccloskey1989catastrophic}
M.~McCloskey and N.~J. Cohen.
\newblock Catastrophic interference in connectionist networks: The sequential
  learning problem.
\newblock {\em Psychology of learning and motivation}, 24:109--165, 1989.

\bibitem{Mensink2012}
T.~Mensink, J.~Verbeek, F.~Perronnin, and G.~Csurka.
\newblock Metric learning for large scale image classification: Generalizing to
  new classes at near-zero cost.
\newblock In {\em European Conference on Computer Vision (ECCV)}, 2012.

\bibitem{mensink2013distance}
T.~Mensink, J.~Verbeek, F.~Perronnin, and G.~Csurka.
\newblock Distance-based image classification: Generalizing to new classes at
  near-zero cost.
\newblock {\em IEEE Transactions on Pattern Analysis and Machine Intelligence
  (T-PAMI)}, 35(11), 2013.

\bibitem{misra2016cross}
I.~Misra, A.~Shrivastava, A.~Gupta, and M.~Hebert.
\newblock Cross-stitch networks for multi-task learning.
\newblock In {\em Conference on Computer Vision and Pattern Recognition
  (CVPR)}, 2016.

\bibitem{misra2014data}
I.~Misra, A.~Shrivastava, and M.~Hebert.
\newblock Data-driven exemplar model selection.
\newblock In {\em Winter Conference on Applications of Computer Vision (WACV)},
  pages 339--346, 2014.

\bibitem{moe2005catastophic}
O.-M. Moe-Helgesen and H.~Stranden.
\newblock Catastophic forgetting in neural networks.
\newblock Technical report, Norwegian University of Science and Technology
  (NTNU), 2005.

\bibitem{MuhlbaierTP09}
M.~D. Muhlbaier, A.~Topalis, and R.~Polikar.
\newblock Learn$^{++}$.{NC}: Combining ensemble of classifiers with dynamically
  weighted consult-and-vote for efficient incremental learning of new classes.
\newblock {\em IEEE Transactions on Neural Networks (T-NN)}, 20(1), 2009.

\bibitem{pentina-cvpr2015}
A.~Pentina, V.~Sharmanska, and C.~H. Lampert.
\newblock Curriculum learning of multiple tasks.
\newblock In {\em Conference on Computer Vision and Pattern Recognition
  (CVPR)}, 2015.

\bibitem{PolikarUUH01}
R.~Polikar, L.~Upda, S.~S. Upda, and V.~Honavar.
\newblock Learn++: an incremental learning algorithm for supervised neural
  networks.
\newblock {\em IEEE Transactions on Systems, Man, and Cybernetics, Part C},
  31(4), 2001.

\bibitem{rannen2017b}
A.~Rannen Triki, R.~Aljundi, M.~B.~Blaschko, and T.~Tuytelaars.
\newblock Encoder based lifelong learning.
\newblock {\em arXiv preprint arXiv:1704.01920}, 2017.

\bibitem{ristin2014incremental}
M.~Ristin, M.~Guillaumin, J.~Gall, and L.~Van~Gool.
\newblock Incremental learning of {NCM} forests for large-scale image
  classification.
\newblock In {\em Conference on Computer Vision and Pattern Recognition
  (CVPR)}, 2014.

\bibitem{robins95}
A.~V. Robins.
\newblock Catastrophic forgetting, rehearsal and pseudorehearsal.
\newblock {\em Connection Science}, 7(2):123--146, 1995.

\bibitem{royer-cvpr2015}
A.~Royer and C.~H. Lampert.
\newblock Classifier adaptation at prediction time.
\newblock In {\em Conference on Computer Vision and Pattern Recognition
  (CVPR)}, 2015.

\bibitem{ILSVRC15}
O.~Russakovsky, J.~Deng, H.~Su, J.~Krause, S.~Satheesh, S.~Ma, Z.~Huang,
  A.~Karpathy, A.~Khosla, M.~Bernstein, A.~C. Berg, and L.~Fei-Fei.
\newblock {ImageNet Large Scale Visual Recognition Challenge}.
\newblock {\em International Journal of Computer Vision (IJCV)}, 115(3), 2015.

\bibitem{rusu2016progressive}
A.~A. Rusu, N.~C. Rabinowitz, G.~Desjardins, H.~Soyer, J.~Kirkpatrick,
  K.~Kavukcuoglu, R.~Pascanu, and R.~Hadsell.
\newblock Progressive neural networks.
\newblock {\em arXiv preprint arXiv:1606.04671}, 2016.

\bibitem{saxena2016convolutional}
S.~Saxena and J.~Verbeek.
\newblock Convolutional neural fabrics.
\newblock In {\em Conference on Neural Information Processing Systems (NIPS)},
  2016.

\bibitem{Scheirer_2013_TPAMI}
W.~J. Scheirer, A.~Rocha, A.~Sapkota, and T.~E. Boult.
\newblock Towards open set recognition.
\newblock {\em IEEE Transactions on Pattern Analysis and Machine Intelligence
  (T-PAMI)}, 36, 2013.

\bibitem{SrivastavaHKSS14}
N.~Srivastava, G.~E. Hinton, A.~Krizhevsky, I.~Sutskever, and R.~Salakhutdinov.
\newblock Dropout: a simple way to prevent neural networks from overfitting.
\newblock {\em Journal of Machine Learning Research (JMLR)}, 15(1), 2014.

\bibitem{Welling09}
M.~Welling.
\newblock Herding dynamical weights to learn.
\newblock In {\em International Conference on Machine Learing (ICML)}, 2009.

\bibitem{xiao2014error}
T.~Xiao, J.~Zhang, K.~Yang, Y.~Peng, and Z.~Zhang.
\newblock Error-driven incremental learning in deep convolutional neural
  network for large-scale image classification.
\newblock In {\em International Conference on Multimedia (ACM MM)}, 2014.

\end{thebibliography}
}

\appendix 

\begin{figure*}\centering
\includegraphics[width=\textwidth]{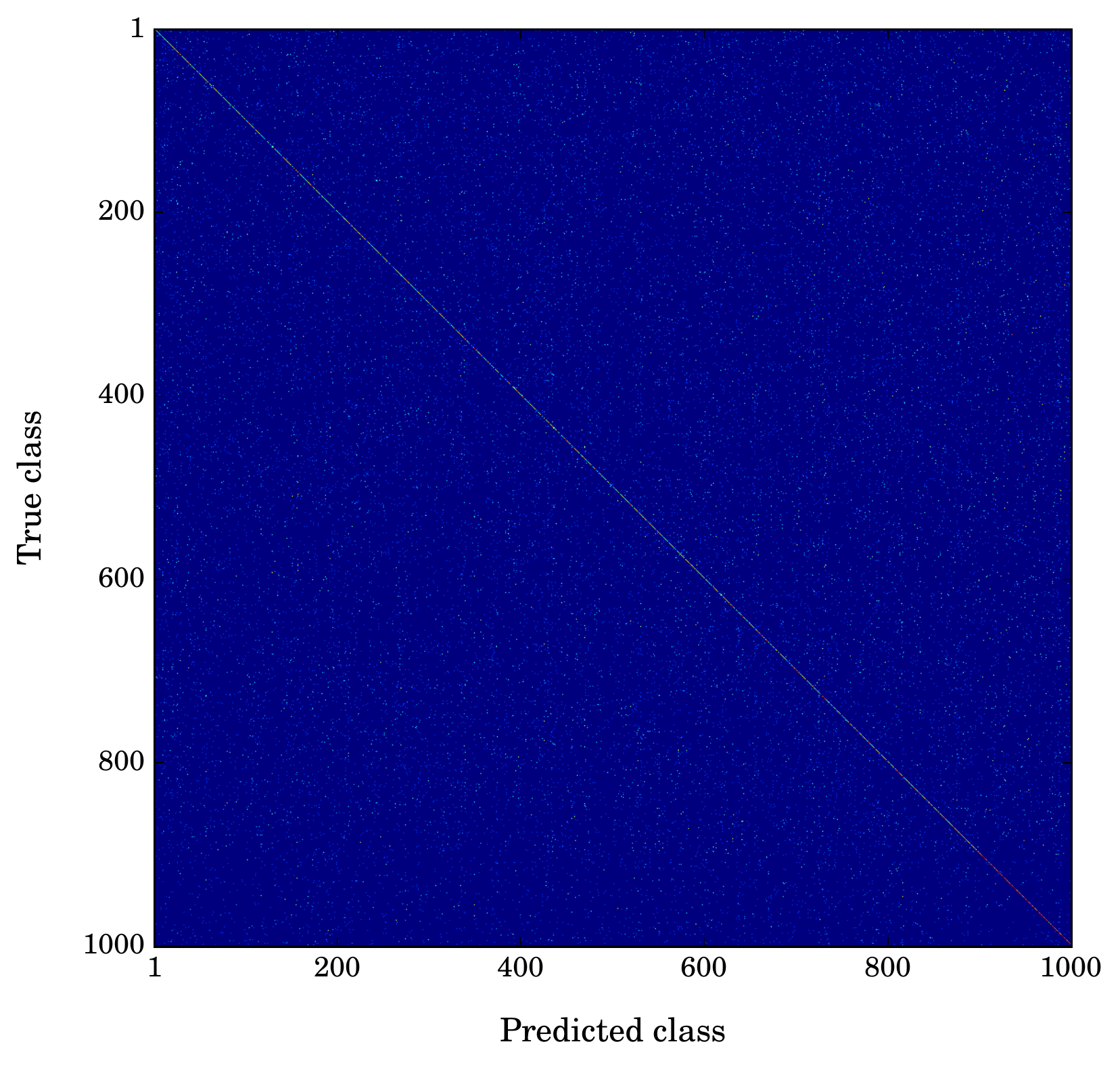}
\caption{Confusion matrix for \emph{iCaRL} on \emph{iILSVRC-large} (1000 classes in batches of 100)}
\label{fig:ILSVRC-confusion1}
\end{figure*}

\begin{figure*}\centering
\includegraphics[width=\textwidth]{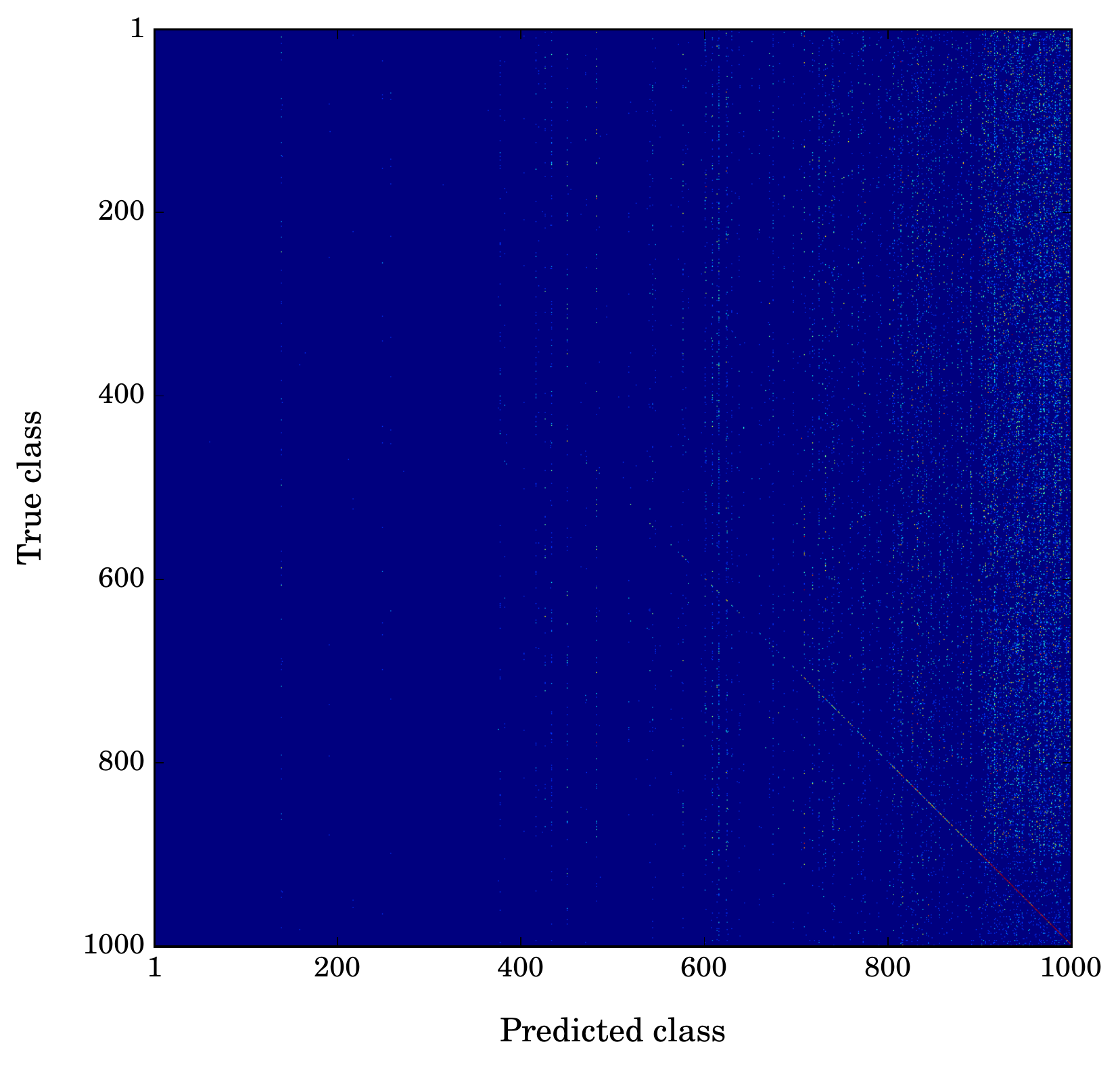}
\caption{Confusion matrix for \emph{LwF.MC} on \emph{iILSVRC-large} (1000 classes in batches of 100)}
\label{fig:ILSVRC-confusion2}
\end{figure*}

\begin{figure*}\centering
\includegraphics[width=\textwidth]{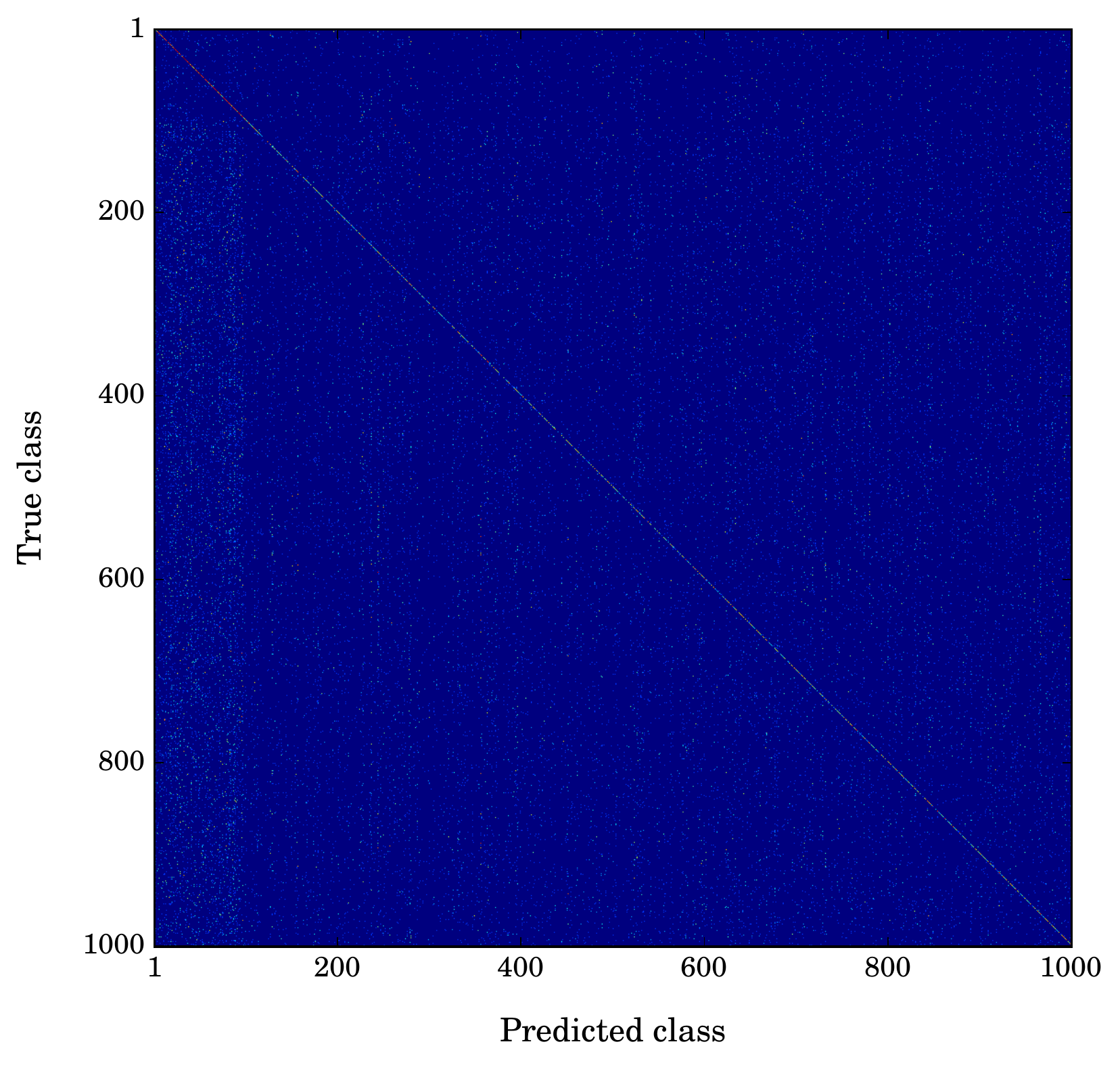}
\caption{Confusion matrix for \emph{fixed representation} on \emph{iILSVRC-large} (1000 classes in batches of 100)}
\label{fig:ILSVRC-confusion3}
\end{figure*}

\begin{figure*}\centering
\includegraphics[width=\textwidth]{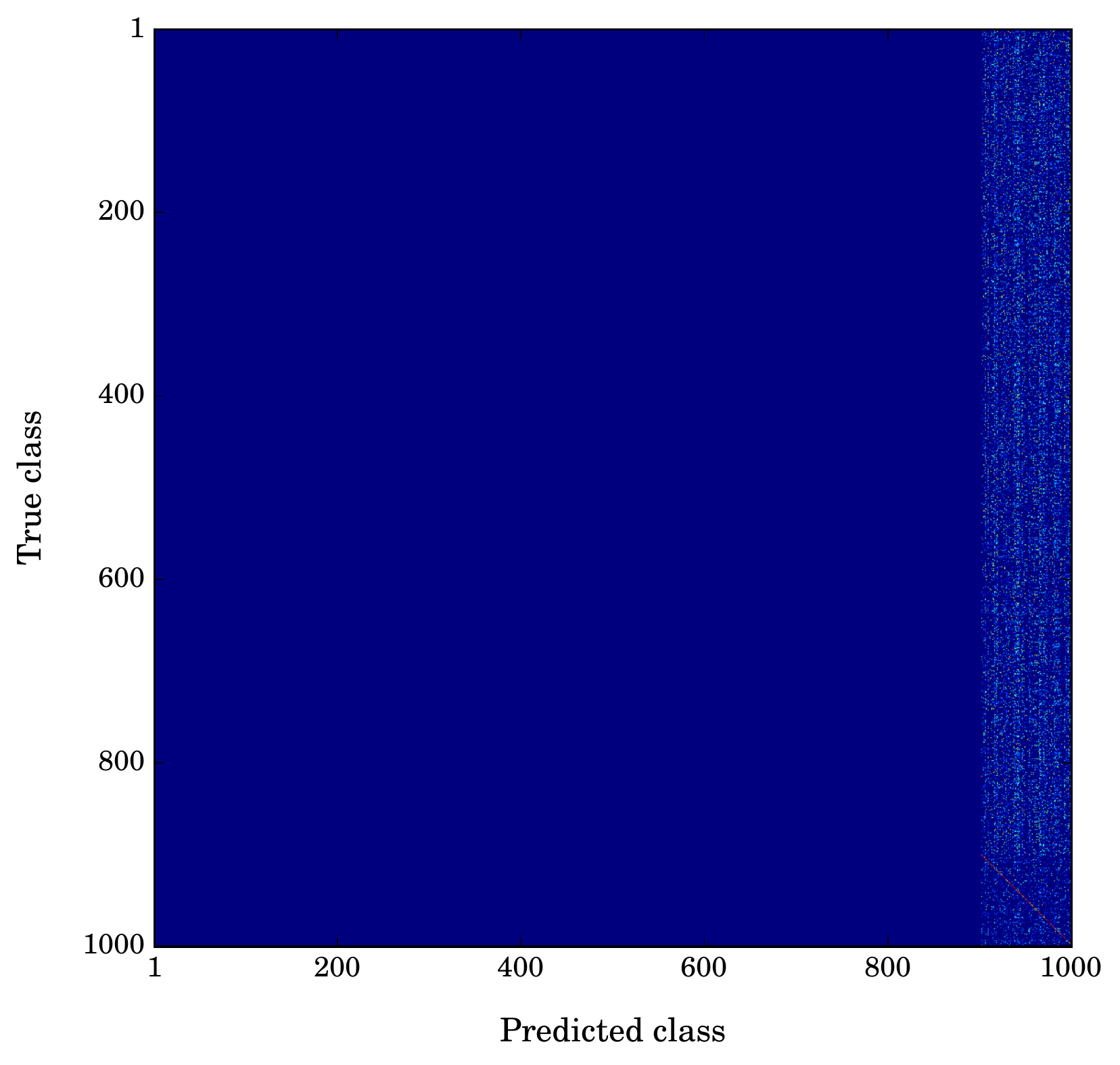}
\caption{Confusion matrix for \emph{finetuning} on \emph{iILSVRC-large} (1000 classes in batches of 100)}
\label{fig:ILSVRC-confusion4}
\end{figure*}

\end{document}